%% file: R2.tex
\documentclass[10pt,journal,compsoc]{IEEEtran}
\usepackage{multirow}
\usepackage[export]{adjustbox}
\usepackage{mathtools}
\usepackage{xcolor}
\usepackage{amsmath}
\usepackage{amssymb}
\usepackage{hyperref}
\usepackage[nolist]{acronym}
\usepackage{subcaption}
\usepackage{soul}
\usepackage[normalem]{ulem}
\ifCLASSOPTIONcompsoc
  \usepackage[nocompress]{cite}
\else
  \usepackage{cite}
\fi
\ifCLASSINFOpdf
\else
\fi

\definecolor{mygc}{RGB}{0,179,80}
\definecolor{myrc}{RGB}{255,0,0}
\newcommand{\myg}{\color{mygc}}
\newcommand{\myr}{\color{myrc}}

\input{sub/_decl}
\input{sub/_acro}

\begin{document}

%
% paper title
% Titles are generally capitalized except for words such as a, an, and, as,
% at, but, by, for, in, nor, of, on, or, the, to and up, which are usually
% not capitalized unless they are the first or last word of the title.
% Linebreaks \\ can be used within to get better formatting as desired.
% Do not put math or special symbols in the title.
\title{Learning to Recognize Actions on Objects in Egocentric Video with Attention Dictionaries}
%\title{Leveraging Attention to Reason about Actions on Objects in Egocentric Video}
%\title{Temporal Reasoning on Spatial Attention for Egocentric Action Recognition}
%\title{Long-Short Term Attention for Egocentric Action Recognition}
%
%
% author names and IEEE memberships
% note positions of commas and nonbreaking spaces ( ~ ) LaTeX will not break
% a structure at a ~ so this keeps an author's name from being broken across
% two lines.
% use \thanks{} to gain access to the first footnote area
% a separate \thanks must be used for each paragraph as LaTeX2e's \thanks
% was not built to handle multiple paragraphs
%
%
%\IEEEcompsocitemizethanks is a special \thanks that produces the bulleted
% lists the Computer Society journals use for "first footnote" author
% affiliations. Use \IEEEcompsocthanksitem which works much like \item
% for each affiliation group. When not in compsoc mode,
% \IEEEcompsocitemizethanks becomes like \thanks and
% \IEEEcompsocthanksitem becomes a line break with idention. This
% facilitates dual compilation, although admittedly the differences in the
% desired content of \author between the different types of papers makes a
% one-size-fits-all approach a daunting prospect. For instance, compsoc 
% journal papers have the author affiliations above the "Manuscript
% received ..."  text while in non-compsoc journals this is reversed. Sigh.

\author{Swathikiran~Sudhakaran, %~\IEEEmembership{Student Member,~IEEE,}
        Sergio~Escalera, %~\IEEEmembership{Member,~IEEE,}
        and~Oswald~Lanz%,~\IEEEmembership{Member,~IEEE}% <-this % stops a space
\IEEEcompsocitemizethanks{
\IEEEcompsocthanksitem S. Sudhakaran is with FBK and University of Trento, Italy. E-mail: sudhakaran@fbk.eu
\IEEEcompsocthanksitem S. Escalera is with Universitat de Barcelona and Computer Vision Center, Spain. E-mail: sergio@maia.ub.es
\IEEEcompsocthanksitem O. Lanz is with FBK, Trento, Italy. E-mail: lanz@fbk.eu}% <-this % stops an unwanted space
%\IEEEcompsocthanksitem O. Lanz is with FBK, Trento, Italy.\protect\\
%E-mail: lanz@fbk.eu}% <-this % stops an unwanted space
%\thanks{Manuscript submitted February 21, revised August 17, 2020.}}%; revised August 26, 2015.}
}

% note the % following the last \IEEEmembership and also \thanks - 
% these prevent an unwanted space from occurring between the last author name
% and the end of the author line. i.e., if you had this:
% 
% \author{....lastname \thanks{...} \thanks{...} }
%                     ^------------^------------^----Do not want these spaces!
%
% a space would be appended to the last name and could cause every name on that
% line to be shifted left slightly. This is one of those "LaTeX things". For
% instance, "\textbf{A} \textbf{B}" will typeset as "A B" not "AB". To get
% "AB" then you have to do: "\textbf{A}\textbf{B}"
% \thanks is no different in this regard, so shield the last } of each \thanks
% that ends a line with a % and do not let a space in before the next \thanks.
% Spaces after \IEEEmembership other than the last one are OK (and needed) as
% you are supposed to have spaces between the names. For what it is worth,
% this is a minor point as most people would not even notice if the said evil
% space somehow managed to creep in.

% The paper headers
\markboth{IEEE Transactions on Pattern Analysis and Machine Intelligence}%
{Shell \MakeLowercase{\textit{et al.}}: Bare Demo of IEEEtran.cls for Computer Society Journals}
% The only time the second header will appear is for the odd numbered pages
% after the title page when using the twoside option.
% 
% *** Note that you probably will NOT want to include the author's ***
% *** name in the headers of peer review papers.                   ***
% You can use \ifCLASSOPTIONpeerreview for conditional compilation here if
% you desire.

% The publisher's ID mark at the bottom of the page is less important with
% Computer Society journal papers as those publications place the marks
% outside of the main text columns and, therefore, unlike regular IEEE
% journals, the available text space is not reduced by their presence.
% If you want to put a publisher's ID mark on the page you can do it like
% this:
\IEEEpubid{\href{https://doi.ieeecomputersociety.org/10.1109/TPAMI.2021.3058649}{10.1109/TPAMI.2021.3058649}~\copyright~2021 IEEE}
% or like this to get the Computer Society new two part style.
%\IEEEpubid{\makebox[\columnwidth]{\hfill 0000--0000/00/\$00.00~\copyright~2015 IEEE}%
%\hspace{\columnsep}\makebox[\columnwidth]{Published by the IEEE Computer Society\hfill}}
% Remember, if you use this you must call \IEEEpubidadjcol in the second
% column for its text to clear the IEEEpubid mark (Computer Society jorunal
% papers don't need this extra clearance.)

% use for special paper notices
%\IEEEspecialpapernotice{(Invited Paper)}

\input{sub/0_abstract}

% for Computer Society papers, we must declare the abstract and index terms
% PRIOR to the title within the \IEEEtitleabstractindextext IEEEtran
% command as these need to go into the title area created by \maketitle.
% As a general rule, do not put math, special symbols or citations
% in the abstract or keywords.
%\IEEEtitleabstractindextext{%
%\begin{abstract}
%\input{sub/0_abstract}
%\end{abstract}

% Note that keywords are not normally used for peerreview papers.
%\begin{IEEEkeywords}

% make the title area
\maketitle

%\IEEEdisplaynontitleabstractindextext
% To allow for easy dual compilation without having to reenter the
% abstract/keywords data, the \IEEEtitleabstractindextext text will
% not be used in maketitle, but will appear (i.e., to be "transported")
% here as \IEEEdisplaynontitleabstractindextext when the compsoc 
% or transmag modes are not selected <OR> if conference mode is selected 
% - because all conference papers position the abstract like regular
% papers do.
\IEEEdisplaynontitleabstractindextext
% \IEEEdisplaynontitleabstractindextext has no effect when using
% compsoc or transmag under a non-conference mode.

% For peer review papers, you can put extra information on the cover
% page as needed:
% \ifCLASSOPTIONpeerreview
% \begin{center} \bfseries EDICS Category: 3-BBND \end{center}
% \fi
%
% For peerreview papers, this IEEEtran command inserts a page break and
% creates the second title. It will be ignored for other modes.
\IEEEpeerreviewmaketitle

\IEEEraisesectionheading{\section{Introduction}\label{sec:introduction}}
\label{sec:intro}
\input{sub/1_intro}

\section{Related Work}
\label{sec:related_work}
\input{sub/2_related}

\section{Action-on-Object Prediction Framework}
\label{sec:method}
\input{sub/3_method}

\section{Experiments}
\label{sec:experiments}

\input{sub/4_results}

\section{Conclusion}
\label{sec:conclusion}
\input{sub/5_conclusions}

\section*{Acknowledgements}
We gratefully acknowledge the support from Amazon AWS Machine Learning Research Awards (MLRA) and NVIDIA AI Technology Center (NVAITC), EMEA. This work has been partially supported by the Spanish project PID2019-105093GB-I00 (MINECO/FEDER, UE), CERCA Programme/Generalitat de Catalunya, and ICREA under the ICREA Academia programme.

\bibliographystyle{ieee}
\bibliography{R2}

% biography section
% 
% If you have an EPS/PDF photo (graphicx package needed) extra braces are
% needed around the contents of the optional argument to biography to prevent
% the LaTeX parser from getting confused when it sees the complicated
% \includegraphics command within an optional argument. (You could create
% your own custom macro containing the \includegraphics command to make things
% simpler here.)
%\begin{IEEEbiography}[{\includegraphics[width=1in,height=1.25in,clip,keepaspectratio]{mshell}}]{Michael Shell}
% or if you just want to reserve a space for a photo:

\newpage
%\vskip -5mm
\begin{IEEEbiography}[{\includegraphics[width=1in,height=1.25in,clip,keepaspectratio]{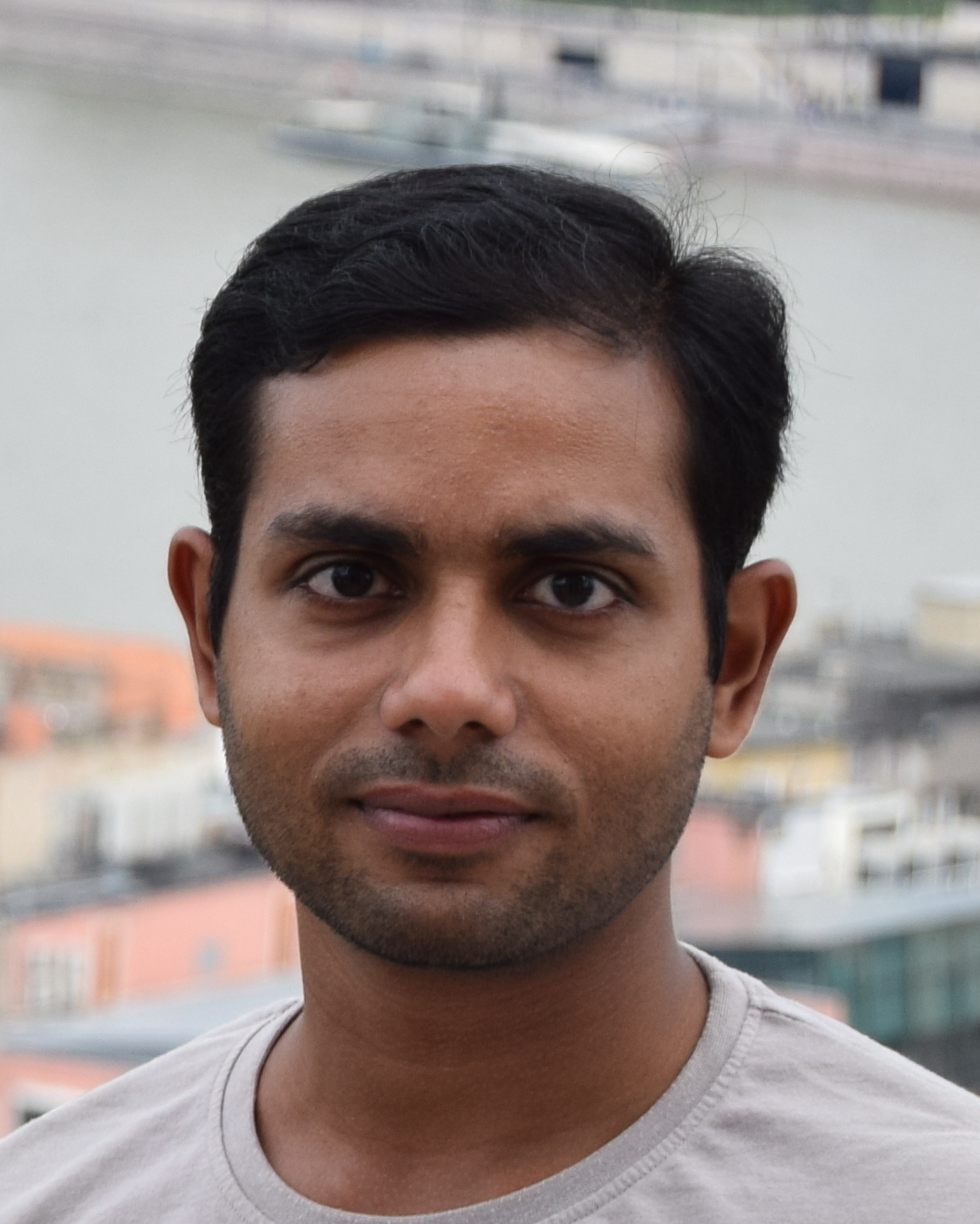}}]{Swathikiran Sudhakaran}
is a post-doctoral researcher at University of Trento, Italy. He received his Ph.D. (Cum Laude) in Computer Science from University of Trento, Italy in 2019. He is also a graduate of Master of Technology from University of Kerala, India. During his Ph.D., he was part of the Technologies of Vision (TeV) lab in FBK, advised by Dr. Oswald Lanz. His research interests are in the area of efficient and effective video representation learning, specifically human behavior understanding.
\end{IEEEbiography}

\begin{IEEEbiography}[{\includegraphics[width=1in,height=1.25in,clip,keepaspectratio]{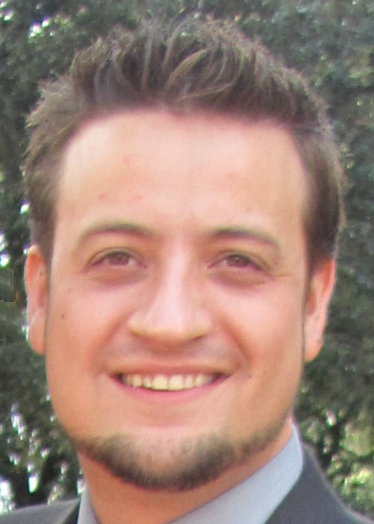}}]{Sergio Escalera}
obtained the 2008 best Thesis award on Computer Science at Universitat Autònoma de Barcelona. He is ICREA Academia and Full Professor at the Department of Mathematics and Informatics, Universitat de Barcelona. He is member of the Computer Vision Center, UAB. He is series editor of The Springer Series on Challenges in Machine Learning. He is vice-president of ChaLearn Challenges in Machine Learning, leading ChaLearn Looking at People events. He is Fellow of the European Laboratory for Learning and Intelligent Systems ELLIS and Chair of IAPR TC-12: Multimedia and visual information systems. He has been General co-Chair of FG20. His research interests include automatic analysis of humans from visual and multi-modal data.
\end{IEEEbiography}

\begin{IEEEbiography}[{\includegraphics[width=1in,height=1.25in,clip,keepaspectratio]{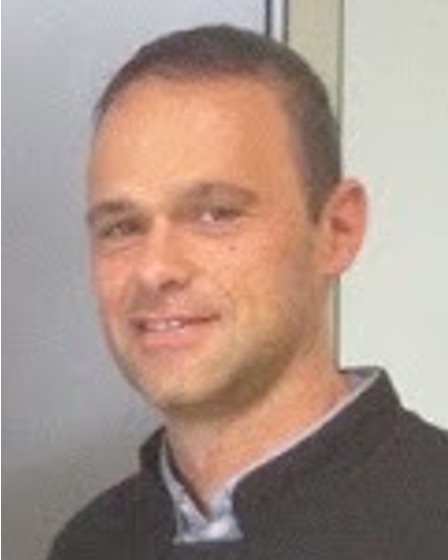}}]{Oswald Lanz}
is Research Scientist at Fondazione Bruno Kessler (FBK), Trento, Italy. He holds the MSc degree in Mathematics and a Ph.D. in Computer Science from the University of Trento. His research belongs to the area of computer vision and machine learning and is centered around video understanding and 3d scene analysis. He has been invited speaker at the ICCV 2019 Workshop on Egocentric Perception, Interaction and Computing (EPIC) and received the Amazon Research Awards (AWS MLRA) for research on representation learning for video in 2020. He is member of the European Lab for Learning and Intelligent Systems ELLIS and was head of the computer vision lab at FBK from 2012 to 2015.
\end{IEEEbiography}
\vfill

% You can push biographies down or up by placing
% a \vfill before or after them. The appropriate
% use of \vfill depends on what kind of text is
% on the last page and whether or not the columns
% are being equalized.

%\vfill

% Can be used to pull up biographies so that the bottom of the last one
% is flush with the other column.
%\enlargethispage{-5in}

% that's all folks
\end{document}

%% file: sub/_decl.tex
\let\tanh\relax

\DeclareMathOperator\tanh{\eta}

\DeclareMathOperator*{\argmax}{arg\,max}
\newcommand{\io}[1]{{\bf #1}}
\def\cnv{*}
\newcommand\ie{\emph{i.e}.}
\newcommand\eg{\emph{e.g}.}
\newcommand\etal{\emph{et al}. }
\newcommand\etc{\emph{etc}.}

\graphicspath{{figs/}}

%% file: sub/_acro.tex
\newacro{clstm}[ConvLSTM]{Convolutional Long Short-Term Memory}
\newacro{lstm}[LSTM]{Long Short-Term Memory}
\newacro{lsta}[LSTA]{Long Short-Term Attention}
\newacro{rnn}[RNN]{Recurrent Neural Network}
\newacro{cnn}[CNN]{Convolutional Neural Network}
\newacro{sgd}[SGD]{Stochastic Gradient Descent}

%% file: sub/0_abstract.tex
% for Computer Society papers, we must declare the abstract and index terms
% PRIOR to the title within the \IEEEtitleabstractindextext IEEEtran
% command as these need to go into the title area created by \maketitle.
% As a general rule, do not put math, special symbols or citations
% in the abstract or keywords.
\IEEEtitleabstractindextext{%
\begin{abstract}
We present EgoACO, a deep neural architecture for video action recognition that learns to pool action-context-object descriptors from frame level features by leveraging the verb-noun structure of action labels in egocentric video datasets. The core component of EgoACO is class activation pooling (CAP), a differentiable pooling operation that combines ideas from bilinear pooling for fine-grained recognition and from feature learning for discriminative localization. CAP uses self-attention with a dictionary of learnable weights to pool from the most relevant feature regions. Through CAP, EgoACO learns to decode object and scene context descriptors from video frame features. For temporal modeling in EgoACO, we design a recurrent version of class activation pooling termed Long Short-Term Attention (LSTA). LSTA extends convolutional gated LSTM with built-in spatial attention and a re-designed output gate. Action, object and context descriptors are fused by a multi-head prediction that accounts for the inter-dependencies between noun-verb-action structured labels in egocentric video datasets. EgoACO features built-in visual explanations, helping learning and interpretation. Results on the two largest egocentric action recognition datasets currently available, EPIC-KITCHENS and EGTEA, show that by explicitly decoding action-context-object descriptors, EgoACO achieves state-of-the-art recognition performance. 
\end{abstract}

% Note that keywords are not normally used for peerreview papers.
\begin{IEEEkeywords}
Egocentric Vision, Action Recognition, Attention, Higher Order Pooling, Video Classification
\end{IEEEkeywords}}

%% file: sub/1_intro.tex
\IEEEPARstart{E}{gocentric Vision}, or first-person vision, concerns the development of computational approaches to visual intelligence for wearable cameras. Such cameras are typically worn on the head or on the chest and naturally approximate the visual field of the camera wearer, thereby capturing people's moments in the way they perceive them visually in natural environment. Making the video information from egocentric perception and computing devices accessible through computer vision and usable via artificial intelligence has huge impact potential in people's life. Some of the applications include tracking and analysis of behavior to regulate unhealthy lifestyle, providing support to elderly and visually impaired, advancing human-robot interaction, studying human-human behavior and cognition, \etc, to name a few.

Perceptual attributes such as observer gaze and spatial location, and 3D object and scene information are more directly measurable or discernible from egocentric camera and sensor streams by matured technology (available on HoloLens2 and MagicLeap, \eg). However, other relevant aspects of interest such as monitoring complex human activities require a higher degree of learning-from-streams capability. Sequences of object manipulation actions are more intrinsically instilled in the sensory signals. They unfold into human activities at a longer temporal scale and are thus harder to characterize visually, identify and measure, thus calling for a learning approach. A fine-grained recognition requires focus on subtle discriminative information that unravels spatially as well as temporally within a visually cluttered, dynamic scene. In egocentric video such information can unveil intermittently in time (during movement of the wearer with abrupt view changes), or be a latent feature of the scene (an open dish cleaner while clearing),
or require spatial-temporal reasoning to be distilled out of a frame sequence (when an object is manipulated under an action).

Action recognition in egocentric video is particularly challenging because of several factors that are peculiar to the scenarios and the way they are recorded in natural setting and for research purpose. Most research is devoted to the analysis of video captured from distant, third-person views, where the actor is usually entirely visible in a static view of the scene. In first-person video, typically one only observes the hands and the objects being manipulated under an action, and their apparent motion is intertwined with scene ego-motion induced by the frequent body movement of the camera wearer. This ego-motion may or may not be representative of the action performed by the observer. As a result, existing techniques proposed for general action recognition become less suitable for egocentric video analysis. Another peculiarity is that there is typically one 'active object' among many in the scene that an action is being applied upon. Which among the many candidate objects in a cluttered scene is going to be the active object depends on temporal context, and temporal reasoning may be required to localize its action specific features spatially across a frame sequence. Also, there is some scene context that can help to predict the active object and the action that is going to be applied on it. Relevant context may come into view at different stages of an action, even in a non causal order, and be thus conveniently sampled at frame level and accumulated over time into an aggregated representation. In contrast, the representation of latent contextual scene features  such as those of a dominant object, will rely on evidence across the video to be detected and extracted.
Representation learning with egocentric action video can therefore benefit from a proper network design that reflects these differences: video descriptors for scene context, active object, and temporal reasoning on it should differ in the way they are extracted spatially from a shared representation and aggregated spatio-temporally. The aim of this paper is to validate this hypothesis and its feasibility for egocentric action recognition. To achieve this we investigate on the use of specialized attention mechanisms as a key element in developing egocentric perception capabilities.

To address the challenge of human action recognition in egocentric vision, in this paper we present a deep neural architecture, EgoACO, whose design roots in the need for explicit action-context-object feature encoding discussed above. The core component of EgoACO is a differentiable pooling operation, termed Class Activation Pooling (CAP), that combines ideas from bilinear pooling for fine-grained recognition and from feature learning for discriminative localization. It uses self-attention with a dictionary of learnable weights to learn to pool from the most relevant feature regions. EgoACO uses CAP to decode object and scene context descriptors from video frame features. It does so by changing the pooling scheme from global (video level, for dominant object) to temporally local (frame level, for scene context). For temporal modeling in EgoACO, we design a recurrent version of class activation pooling termed Long Short-Term Attention (LSTA) which is our core technical contribution. LSTA extends convolutional gated LSTM with built-in spatial attention and a re-designed output gate. It is a general model that can be used in RNN based sequence learning architectures, while it is instantiated in EgoACO for temporal encoding of the action descriptor. Action, object and context descriptors are finally fused by a multi-head prediction that re-enforces the inter-dependencies between noun-verb-action structured labels in egocentric action recognition datasets. We provide an extensive model analysis and ablation study of EgoACO and comparison to state of the art using the two largest egocentric action recognition datasets currently available, EPIC-KITCHENS and EGTEA Gaze+. Furthermore, we show through examples that predictions are consistent with generated attention maps. Indeed, EgoACO features built-in visual explanations, helping the learning and interpretation of discriminative information in video. It also shows that the relevant feature regions are not necessarily box shaped or confined to video tubes, which are representations commonly used for localization tasks.

The paper is organized as follows. Section~\ref{sec:related_work} discusses work in egocentric action recognition most closely related to ours. In Section~\ref{sec:method}, we present the technical details of the EgoACO framework. We present and discuss experimental results in Section~\ref{sec:experiments} and draw conclusions in Section~\ref{sec:conclusion}. 

%% file: sub/2_related.tex
In videos, an action is governed by two types of information, appearance and motion. A successful action recognition approach should be able to encode information across the two aforementioned domains. Appearance cues are characterized by the challenges present in image recognition such as variations in view point and environment. Moreover, the flexible nature of human body further adds variations in the action being carried out among different subjects. On the other hand, actions can be carried out at different temporal speed, which complicates the temporal encoding of appearance cues. 

With the increasing success of deep learning in addressing image classification problems~\cite{resnet, resnext, inception}, several deep learning based approaches have been developed for video action recognition. The straight-forward approach is to extend 2D~\acp{cnn} to the temporal dimension and applying stacks of video frames as the input~\cite{3dcnn, tran3dcnn, hara2018can}, thereby generating spatio-temporal features. Another alternative is to extract appearance features from the individual frames and perform a temporal pooling operation to encode their temporal evolution~\cite{lrcn, tsn, fernando2016rank}. Recent approaches explore the feasibility of temporal modeling with 2D CNNs~\cite{r2+1d, tsm, Sudhakaran_2020_CVPR}. 
Another approach includes using two \acp{cnn}, each encoding an RGB image for appearance cues and stacks of optical flow for motion cues~\cite{simonyan2014two, feichtenhofer2016convolutional, christoph2016spatiotemporal}.

Compared to third-person videos, egocentric videos are further challenging due to its intrinsic properties. Firstly, egocentric videos lack the body pose of the actor and have to rely on the hand motion and/or ego-motion. Secondly, the ego-motion caused by the camera wearer may or may not be representative of the action. This can result in erroneous predictions. Thirdly, the presence of cluttered background with several objects can lead to the model learning wrong and suboptimal representations.

\subsection{First Person Action Recognition}

Action recognition approaches developed for third-person videos~\cite{simonyan2014two, tsn, carreira2017quo} may be ineffective for egocentric videos because of previous discussed challenges. However, objects and hands contained in egocentric videos provide relevant information that is not available in third-person videos. Several works that utilize such information have been proposed for egocentric activity recognition. \cite{baradel2018object}, \cite{aboubakr2019recognizing} propose to use information about the objects present in the scene to recognize activities. The works of \cite{ma2016deeper, singh2016first, zhou2016cascaded, h+o} train specialized CNNs for hand segmentation and object localization related to the activities to be recognized. These methods base on specialized pre-training for hand segmentation and object detection networks, requiring high amounts of annotated data for that purpose. Gaze information present in egocentric videos is also found to be useful for activity recognition~\cite{li2015delving, huang2019mutual}.

Another line of research consists of developing effective techniques for extracting temporal information present in the video.  In \cite{ryoo2015pooled, zaki2017modeling} features are extracted from a series of frames to perform temporal pooling with different operations, including max pooling, sum pooling, or histogram of gradients. Then, a temporal pyramid structure allows the encoding of both long term and short term characteristics. However, all these methods do not take into consideration the temporal order of the frames. Techniques that use a recurrent neural network such as \acf{lstm}~\cite{cao2017egocentric, verma2018making, rulstm}  and its convolutional variant (\acs{clstm})~\cite{sudhakaran2017convolutional, egornn} are proposed to encode the temporal order of features extracted from a sequence of frames.

Adding optical flow extracted from video frames as an additional modality to RGB frames have benefited third person action recognition techniques~\cite{feichtenhofer2016convolutional, simonyan2014two}. Inspired from this, approaches that apply additional modalities of information have been explored in egocentric domain~\cite{ma2016deeper, tang2017action, tang2018multi}. Due to the presence of ego-motion, which may or may not be correlated with the action, directly applying the optical flow can lead to wrong predictions. \cite{ma2016deeper} address this problem by using warped optical flow that compensates the ego-motion present in the input video. \cite{tang2017action, tang2018multi} add an additional stream that accepts depth maps to the two stream network enabling it to encode 3D information present in the scene. Audio present in the videos is also found to be an effective source of information for egocentric action recognition~\cite{tbn, cartas2019seeing}.

Majority of the state-of-the-art techniques rely on additional annotations such as hand segmentation, object bounding box or gaze information~\cite{ma2016deeper, li2018eye, lu2019deep, zhou2016cascaded, h+o}. This allows the network to concentrate on the relevant regions in the frame and helps in distinguishing each activity from one another. However, manually annotating all the frames of a video with previous information is impractical. For this reason, the development of techniques that can identify relevant regions of a frame without using additional annotations is crucial.

\subsection{Attention}

Attention mechanism was proposed to enable a network to focus on features derived from spatial regions that are relevant for the recognition of a given task. This includes \cite{anderson2018bottom, ma2017attend, wang2018bidirectional} for image and video captioning, \cite{nguyen2018improved, anderson2018bottom, liang2018focal} for visual question answering and \cite{tavlad, girdhar2017attentional, li2018videolstm} for third-person action recognition.
\cite{girdhar2017attentional} and~\cite{tavlad}  generate top-down attention that focuses on the spatially important regions for third-person action recognition. Unlike its image counterparts, attention maps generated in videos have to be temporally coherent. Several approaches that take into consideration this temporal aspect have been proposed~\cite{li2018videolstm, zhang2019eleatt, sharma2015action, zhu2019redundancy}. \cite{sharma2015action} and~\cite{zhang2019eleatt} derive the attention map for each frame from the hidden state of a \ac{lstm}. The effectiveness of convolutional structures at the different control gates of a \ac{clstm} and attention is studied in~\cite{zhu2019redundancy}. \cite{li2018videolstm} develop a \ac{clstm} module with in-built spatio-temporal attention.

 Due to the presence of cluttered background, identifying the relevant spatial regions is crucial for egocentric action recognition. The works of \cite{li2018eye, lu2019learning, lu2019deep} exploit gaze information present in the videos to generate the attention map. The work of \cite{zhang2019self} uses attention derived from an object detector and temporal cues from a 3D~\ac{cnn} for identifying the objects of interest in egocentric videos. In~\cite{egornn} top-down attention generated from the prior information encoded in a CNN pre-trained for object recognition is used for egocentric action recognition. \cite{matsuo2014attention, sap} use object detectors for generating attention.

Most existing techniques for generating spatial attention in egocentric videos rely on additional information such as eye gaze or object detections. This makes such approaches impractical. Moreover, in these approaches the spatial attention is derived independently for each frame. Since video frame sequences have an absolute temporal consistency, per frame processing results in the loss of valuable information. Moreover, approaches that rely on eye gaze or object detections may neglect another relevant appearance cue present in the video such as scene context.

To address previous issues, in this work we develop a learnable attention mechanism that can leverage the prior information encoded in a \ac{cnn} trained for other tasks, such as object recognition. We further introduce a novel recurrent neural unit, \acf{lsta}, incorporating our attention mechanism. 

\subsection{Multi-task learning}
\label{sec:multitask}

Training a model with multiple objectives, \ie, multi-task learning, is found to be effective in several research areas from speech recognition~\cite{mt-speech1, mt-speech2}, natural language processing~\cite{mt-nlp1, mt-nlp2} to computer vision~\cite{liu2019end, kendall2018multi}. Multi-task learning allows a model to leverage the relation among the considered tasks to generate effective representations, thereby improving its generalization capability.

Several works have explored multi-task learning in the context of egocentric action recognition~\cite{ma2016deeper, kapidis2019multitask, tbn, li2018eye, chalasani2019simultaneous}. \cite{li2018eye} train a network to predict the action along with the gaze while~\cite{kapidis2019multitask} adds hand location prediction as well. Predicting the hand segmentation and gesture jointly is explored in~\cite{chalasani2019simultaneous}. However, these approaches require additional annotations in the form of gaze and hand locations. Other alternative is to exploit the inherent structure of the action labels. In egocentric datasets consisting of object manipulation actions, an action is defined as a verb-noun pair. \cite{tbn} train a network to predict the verb and noun labels while~\cite{ma2016deeper} predicts the action label in addition to the verb and noun labels.

In most of the existing works, the different tasks are predicted from the same representation, usually using a separate output layer~\cite{tbn, li2018eye}. This is less optimal as different tasks may require specialized representations. For instance, verb prediction require strong spatio-temporal reasoning while only spatial reasoning may be required for predicting noun labels. Thus, it is critical to decode the underlying specialized features from frame level representation learned with a \ac{cnn}. Authors of~\cite{ma2016deeper} address this problem by training two different networks, one for verb and one for noun and the output of the two networks is used to predict the action label. Such a schema consisting of different networks is computationally more expensive and difficult to train. In addition, they rely on identical architectures for learning both verb and noun representations.

Here we present EgoACO, an end-to-end trainable deep architecture for egocentric action recognition, exploiting the availability of structured labels and their latent relationship with each other. Our method consists of different independent pooling layers with an in-built attention mechanism. With these specialized layers, our model is capable of decoding the important feature components for addressing each task, \ie, verb, noun and action prediction.

%% file: sub/3_method.tex
\begin{figure}[t!]
    \centering
    \includegraphics[width=\columnwidth]{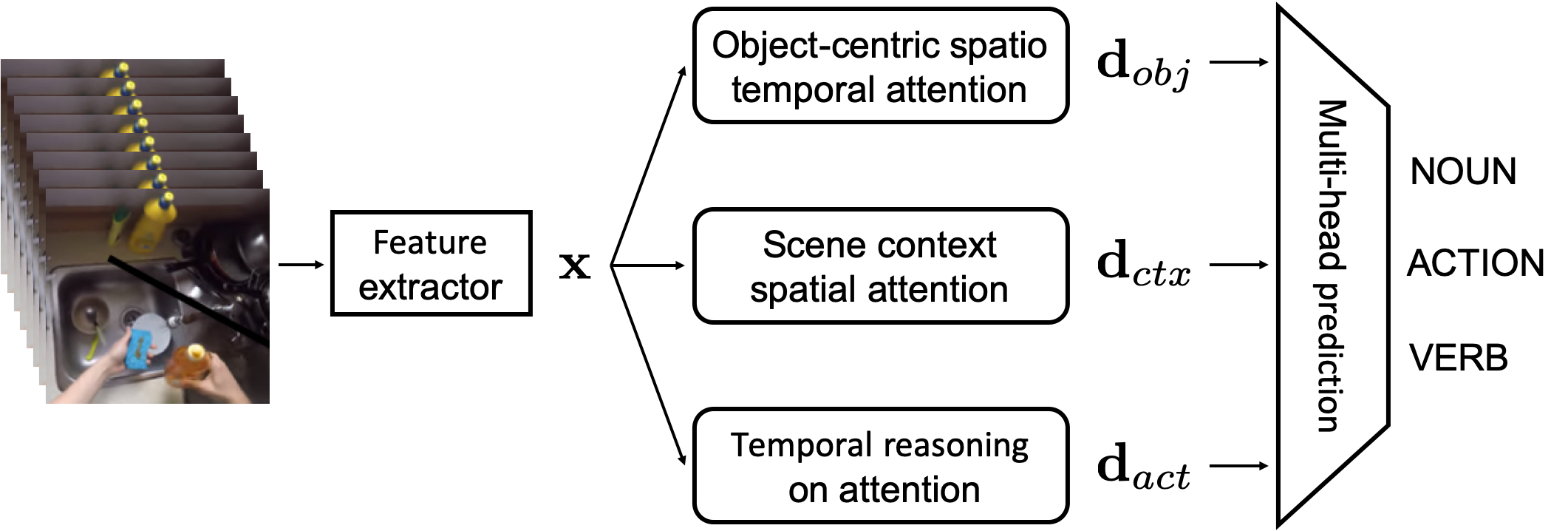}
    \caption{Overview of the egocentric action prediction framework \textbf{EgoACO} -- \textbf{Ego}centric \textbf{A}ction-\textbf{C}ontext-\textbf{O}bject Net.}
    \label{fig:overview}
\end{figure}

In this section, we present a detailed description of our action prediction framework for egocentric video.

\subsection{Overview}
Our model design is inspired by the observation that discriminative information in egocentric video is both appearance based and temporal, and may relate to scene context, a particular object or latent scene feature, and the temporal progression of a performed action. Fig.~\ref{fig:overview} shows an overview of \textbf{Ego}centric \textbf{A}ction-\textbf{C}ontext-\textbf{O}bject Net (\textbf{EgoACO}), our action prediction framework for egocentric video. It is composed of a generic feature extractor and three specialized branches, which are designed to extract descriptors associated to these three factors from the shared video features. The descriptors are finally fused to predict the verb, noun and action categories.

More in detail, given an input video clip, we first sample a temporally ordered sequence of $T$ video frame features $\mathbf{x}$ using a 2D CNN trunk from the ResNet family. From these features we extract three video descriptors, $\io{d}_{obj}, \io{d}_{ctx}, \io{d}_{act}$, to encode the relevant spatial and spatio-temporal information for action prediction in egocentric video. The descriptor extraction blocks implement different {\bf frame feature aggregation schemes} sharing an underlying attention mechanism we term {\bf class activation pooling}, to separate {\bf object-centric, contextual, and temporal reasoning} on the video clip.  The three descriptors are then fused by a {\bf multi-head prediction module} whose design is to explicitly model the inter-dependencies between noun, verb, and action class scores.

In the following we will first introduce Class Activation Pooling (CAP) in subsection \ref{ssec:cap} without making explicit reference to its use in our action prediction architecture. Then, for a smoother and progressive approach to presenting the technical details of EgoACO design, in subsection \ref{ssec:objctx} we first describe how CAP is used to extract object-centric and scene context descriptors from a video clip. How to use it to implement temporal reasoning over a frame feature sequence will then be detailed in subsection \ref{ssec:lsta}, where we introduce LSTA which is our key technical contribution in this paper. We complete the description of our framework with subsection~\ref{ssec:multi} on multi-head prediction.

\subsection{Class Activation Pooling (CAP)}
\label{ssec:cap}

The driving principle of attention design is to make our encoding of clip descriptors effective in the cluttered egocentric scene. We posit this requires the ability to (i) spatially localize a number of descriptor-specific feature regions for objects, context, and temporal reasoning, and (ii) higher order pooling operations to decide upon which of the localized region(s) to encode into the descriptors. To realize this we combine ideas from class activation mapping for discriminative localization~\cite{zhou15cnnlocalization} and bilinear pooling for fine-grained recognition~\cite{lin2015bilinear}.

Let $X \in R^{N \times K}$ be the matrix view of a feature tensor with $N=HW$ spatial locations and $K$ feature channels. First order pooling on $X$ is commonly implemented as the last layer of image classification architectures. By applying a weight matrix $W \in R^{K \times C}$ on the spatially accumulated feature $\mathbf{1}^T X \in R^{1 \times K}$ ($\mathbf{1}$ is a $N$-size vector of all ones), the $C$ class scores are obtained as 
\begin{equation}
pool_{order1}(X) = \mathbf{1}^T X W.\label{eq:pool}
\end{equation}

More generally, a parametric pooling operation will map $X$ to an embedding space of dimension $C$. In second-order parametric pooling, a weight tensor $W \in R^{K \times K \times C}$ is applied to the bilinearly mapped feature $X^T X \in R^{K \times K}$. Each value of the $C$-dimensional embedding can be mathematically expressed as $Tr(X^TXW_c^T)$ where $W_c$ is the $c$-th matrix slice of weight tensor $W$ and $Tr$ is the matrix trace. Since the bilinearly mapped feature is $K^2$-sized hence high-dimensional, full parametric second order pooling is over parametrized and expensive to compute. Low-dimensional approximations such as compact bilinear pooling~\cite{compactbilinear} or attentional pooling~\cite{girdhar2017attentional} have been proposed for vision tasks. Girdhar and Ramanan~\cite{girdhar2017attentional} show that with a rank-1 approximation $W_c \approx \io{a} \io{b}^T$ where $\io{a},\io{b} \in R^{K \times 1}$, the embedding value for dimension $c$ becomes $\io{a}^T X^T X \io{b}$. By imposing that weights $\io{a}$ are shared across embedding dimensions we can stack the $\io{b}$'s into a weight matrix $B \in R^{K \times C}$ and define our rank-1 approximation of second order pooling in matrix notation as
\begin{equation}
    pool_{rank1}(X) = (X\io{a})^T X B.\label{eq:rank}
\end{equation}

We now compare Eq.~\eqref{eq:rank} with standard pooling in Eq.~\eqref{eq:pool} to justify the shared weight assumption on $\io{a}$ and review the attention mechanism in the rank-1 approximation to localize feature regions for second order pooling. A first observation is that $B$ takes on the role of classifier weights $W$ in Eq.~\eqref{eq:pool}. Differently, the spatial sum-pooling operation on $X$ represented by the constant $\io{1}^T$ in Eq.~\eqref{eq:pool} is replaced by a spatial self-attention $(X\io{a})^T \in R^{N \times 1}$. The self-attention acts upon $XB$ by weighting feature locations data dependently\footnote{The shared weight assumption leading to Eq.~\eqref{eq:rank} ensures that a same $(X\io{a})^T$ is to be applied bottom-up across all pooling dimensions.}. 

In a model built on Eq.~\eqref{eq:rank}, the self-attention weights $\io{a}$ is a single, free parameter vector that is learnt during training. For fine-grained recognition in a cluttered scene however, the intuition for a more flexible model is to retrieve $\io{a}$ from a dictionary of learnable self-attention weights. Each weights in the dictionary can then specialize for a specific category. For example, there can be dictionaries for egocentric views of objects, for scene context features, for hand-object views in action. Interestingly, we find that class activation maps~\cite{zhou15cnnlocalization} provide a localization mechanism of the same $(X\io{a})^T$ linear form of Eq.~\eqref{eq:rank} and can be repurposed to implement a weights dictionary extension in rank-1 pooling.

In our extension, a selector maps $X$ into a dictionary score space $\mathcal{A}$ and returns the index $c^*$ of dictionary item that obtained the highest score. The selector is of the form $c^* = \argmax_c\pi(\epsilon(X),\io{a}_c)$ where $\epsilon$ is a reduction and $\io{a}_c \in A$ are the parameters for scoring $X$ against item $c$. If $\pi$ is $\epsilon$-equivariant\footnote{A network layer $\Phi$ is said $g$-equivariant if "transforming an input $x$ by a transformation $g$
(forming $T_g x$) and then passing it through the learned map
$\Phi$ should give the same result as first mapping $x$ through $\Phi$
and then transforming the representation", from Cohen \& Welling: \emph{Group Equivariant Convolutional Networks}, ICML'16.} then $\pi(\epsilon(X), \io{a}) = \epsilon(\pi(X, \io{a}))$ and we can use $\{\epsilon^{\perp}(\pi(\cdot,\io{a}_c))\}_{\io{a}_c \in A}$ as a dictionary of self-attention functions associated to $\epsilon$. That is, self-attention $\varsigma$ from dictionary $A$ is the triplet
\begin{equation}
(\varsigma) = (\epsilon, \pi, A)\text{ , }\quad\pi\textrm{ is }\epsilon\text{-equivariant}
\end{equation}
and is evaluated on feature matrix $X$ by
\begin{eqnarray}
\varsigma(X,A) &=& \epsilon^{\perp}(\pi(X,\io{a}_{c^*}))\label{eq:pooling.0} \label{eq:attention}\\
\textrm{where }c^* &=& \argmax_c\pi(\epsilon(X),\io{a}_c)\label{eq:pooling.1}
\end{eqnarray}
where $\epsilon^{\perp}$ denotes the $\epsilon$-orthogonal reduction\footnote{For example, if $\epsilon$ is max or mean pooling along one dimension then $\epsilon^{\perp}$ is max or mean pooling along the other dimensions.}. If we choose
\begin{eqnarray*}
	\epsilon(X) 
	&\leftarrow& \text{ spatial sum-pooling}\\
	\pi(\epsilon(X),\io{a}_c) 
	&\leftarrow& \text{ linear mapping}
\end{eqnarray*}
then $\varsigma(X,A)XB = (X\io{a}_{c^*})^T XB$ with $c^*$ from Eq.~\eqref{eq:pooling.1}. It has the same linear form as in Eq.~\eqref{eq:rank}. This is related to class activation mapping~\cite{zhou15cnnlocalization} introduced for discriminative localization. Note however that, in contrast to~\cite{zhou15cnnlocalization} that uses strong supervision to train the selector directly, we leverage video-level annotation to implicitly learn an attention mechanism for video classification.

In summary, {\bf class activation pooling} is a differentiable rank-1 approximation of second order pooling with learnable embedding matrix $B$ and self-attention dictionary $A$. The resulting model is
\begin{eqnarray}
    ca\text{-}pool(X, A) &=& softmax(X\io{a}_{c^*})^T X B\label{eq:cap.0}\\
    \textrm{where }c^* &=& \argmax_c \io{1}^T X \io{a}_c\label{eq:cap.1}
\end{eqnarray}
It enhances the model in Eq.~\eqref{eq:rank} by replacing self-attention with class activation based attention to leverage discriminative localization for descriptor encoding. It has the form of a bilinear model $ca\text{-}pool(X, A) = f_A(X)^T f_B(X)$ introduced in~\cite{lin2015bilinear} for fine-grained recognition tasks, where $f_B(X) = XB$ and $f_A(X) = softmax(X\io{a}_{c^*})$ can be interpreted as a detector of discriminative feature regions.

%%%%%%%%%%%%%%%%%%%%%%%%%%%%%%%%%%%%%%%%%%%%%%%%%%%%%%%%%
\subsection{Active Object and Scene Context Pooling}
\label{ssec:objctx}

We now describe how class activation pooling can be used in the architecture of Fig.~\ref{fig:overview} to extract object-centric and scene context descriptors from a video clip.

Action prediction in our context requires identifying the {\bf active object}, which is the object on which the action is applied on. Egocentric scenes are typically cluttered, meaning there can be many objects visible in a first person view of the scene while the wearer is doing an action. Some of these objects may represent {\bf scene context} helpful in recognizing an action, while the presence of others may stay in no relation to the action. Using off-the-shelf object detectors trained on images or single video frames may not be the optimal choice, as it will fire lots of false positives across an egocentric video frame sequence. Also, discriminating the active object from others in the scene may require focus on subtle visual interactions like the contact of a hand with the pizza dough or the spoon entering a sugar bowl. These may not be well represented in object detection datasets and appear only in a fraction of frames in the clip. Densely annotating video clips at frame level with bounding boxes or accurate pixel level masks is difficult and too expensive. We therefore follow a weakly supervised approach to active object and scene context representation learning from video using video-level labels. To face the challenges of fine-grained recognition, we build upon the strong feature learning and discriminative localization capability with class activation pooling. 

\subsubsection{Active Object Descriptor}
We use spatio-temporal class activation pooling over the video feature volume. For this, the tensor $\io{x} \in R^{T \times H \times W \times K}$ is reshaped into its matrix view, $X\in R^{HWT\times K}$. We obtain $\io{d}_{obj} \in R^{M_{obj}}$ using an {\bf object attention dictionary} $A_{obj} \in R^{K \times C}$.
\begin{eqnarray}
    X &\leftarrow& \verb+reshape+(\io{x}; (HWT,K))\label{eq:obj.0}\\
    \io{d}_{obj} &=& ca\text{-}pool(X, A_{obj})
\end{eqnarray}
The active object dictionary $A_{obj}$ will be trained through $\io{d}_{obj}$ receiving direct supervision from the noun category labels. However, through multi-head prediction described in Sec.~\ref{ssec:multi}, both action and verb category labels will also contribute.

\subsubsection{Scene Context Descriptor}
We use frame-wise, spatial class activation pooling over the temporal sequence of frame features. For this, the feature tensor $\io{x} \in R^{T \times H \times W \times K}$ is reshaped into a batch $X$ of $T$ matrix views{\color{blue},} $X \in R^{T \times HW \times K}$. We obtain a batch of frame descriptors using a {\bf context attention dictionary} $A_{ctx} \in R^{K \times C}$. We reduce the batch of $T$ video frame descriptors to $\io{d}_{ctx} \in R^{M_{ctx}}$ using sum pooling. 
\begin{eqnarray}
    X &\leftarrow& \verb+reshape+(\io{x}; (T,HW,K))\\
    \io{d}_{ctx} &=& \io{1}^T ca\text{-}pool(X, A_{ctx})\label{eq:ctx.1}
\end{eqnarray}
The scene context dictionary $A_{ctx}$ will be trained through $\io{d}_{ctx}$ receiving direct supervision from the action category labels, but also verb and noun labels will contribute through coupled multi-head prediction.

%%%%%%%%%%%%%%%%%%%%%%%%%%%%%%%%%%%%%%%%%%%%%%%%%%%%%%%%%
\subsection{Long Short-Term Attention (LSTA)}
\label{ssec:lsta}

We now present LSTA, a new recurrent unit blending the attention mechanism and sequence learning for temporal reasoning over a frame features sequence.

Class activation pooling entails a sophisticated attention mechanism while it uses a linear model, $B$ in Eq.~\eqref{eq:cap.0}, for embedding. We used spatio-temporal reshaping in Eq.~\eqref{eq:obj.0} or sum pooling across time in Eq.~\eqref{eq:ctx.1} to handle the temporal dimension in the video clip. With these operations, video frames are processed as a {\bf set of features}, in an {\bf order-less} manner. Recognizing actions on objects, however, also requires strong {\bf temporal sequential reasoning}. We therefore now replace the linear $B$ in class activation pooling with time-recurrent convolutional embedding. For this we isolate the attention part of class activation pooling in Eq.~\eqref{eq:cap.0} as
\begin{equation}
     ca\text{-}attn(X, A) = softmax(X\io{a}_{c^*})^T \odot X\label{eq:attn}
\end{equation}
where $\odot$ denotes the column-wise Hadamard product between a row vector and a matrix. Since this function outputs a matrix view of the same shape of input $X$, it is licit to write $ca\text{-}attn(\io{x}, A)$ in tensor form, leaving intended the reshape operations to be performed on input $\io{x}$ and its output.

\subsubsection{Action Descriptor Encoding with Recurrent Attention}
\label{sssec:actenc}
We use LSTM~\cite{gers,7508408} with convolutional gates~\cite{NIPS2015_5955} as our recurrent model base. The straightforward way in~\cite{egornn} is to feed the ConvLSTM with attention-weighted frame features sequence, that is, with $ca\text{-}attn(\io{x}, A_{act})$ where $A_{act} \in R^{K \times C}$ is an {\bf action features attention dictionary}. However, videos have an absolute temporal consistency, and the same may hold for the spatial location of discriminative action features. They should track smoothly across an action sequence. Generating spatial attention independently for each frame, that is bottom-up without temporal causal interaction, may result in poor action descriptor encoding. 

We reformulate Eq.~\eqref{eq:attn} to inject temporal dependencies through a residual correction $\io{r}$. We use a convolutional RNN\footnote{Our implementation has LSTM with 1 output plane} to learn the sequence of residuals to have a smooth tracking of attention, that is
\begin{eqnarray}
 rca\text{-}attn(\io{x_t}, \io{r_{t-1}}, A_{act}) &\!\!\!=\!\!\!& 
 \io{r_t}, \io{y_t}\label{eq:lsta.attn.0}\\
 \text{with } \io{r_t} &\!\!\!=\!\!\!& convrnn(\io{x_t}\io{a}_{c^*}; \io{r_{t-1}})\label{eq:lsta.attn.1}\\
 \io{y_t} &\!\!\!=\!\!\!& softmax(\io{x_t}\io{a}_{c^*} \!+ \io{r_t})^T \!\odot \io{x_t}\phantom{x,x}\label{eq:lsta.attn.2}
\end{eqnarray}
To encode the video frame features sequence $\io{x}$ of length $T$ into an action descriptor $\io{d}_{act}$ we apply a single layer \ac{clstm} with $M_{act}=512$ output planes on the output of $rca\text{-}attn$. That is,
\begin{eqnarray}
    \io{r_t}, \io{y_t} &=& rca\text{-}attn(\io{x_t}, \io{r_{t-1}}, A_{act})\label{eq:lsta:attn.0}\\
    \io{h_t},\io{c_t} &=& convlstm(\io{y_t}; \io{h_{t-1}},\io{c_{t-1}})\\
    \io{d}_{act} &=& \io{1}^T \io{c}_T
\end{eqnarray}
where $\io{r}, \io{h}, \io{c}$ are the output and memory state sequences of the RNNs. After recurrent encoding, the final internal memory tensor $\io{c}_T$ is spatial sum-pooled to obtain the action descriptor $\io{d}_{act}$.

\subsubsection{Improving Memory Propagation through Attention}
\label{sec:output_pooling}
We now dive into the internals of $convlstm$ to improve its memory propagation through attention. In compact notation, LSTM equations are
\begin{eqnarray}
(i_c,f_c,c) &\!\!\!\!=\!\!\!\!& (\sigma,\sigma,\tanh)(W_c \cnv [\io{y_t}, \io{o_{t-1}} \odot \tanh(\io{c_{t-1}})])\label{eq:lstm.0}\\
\io{c_t} &\!\!\!\!=\!\!\!\!& f_c \odot \io{c_{t-1}} + i_c \odot c\label{eq:lstm.1}\\
\io{o_t} &\!\!\!\!=\!\!\!\!& \sigma(W_o * [\io{y_t}, \io{o_{t-1}} \odot \tanh(\io{c_{t-1}})])\label{eq:lstm.2}
\end{eqnarray}
where $\sigma,\eta$ are sigmoid and tanh activations, $W_c,W_o$ are the learnable convolution kernels, $*$ is convolution. For convenience, we replace $\io{h_t}$ by $\io{h_t} \leftarrow \io{o_t} \odot \eta(\io{c_t})$ and recur on the output gate $\io{o_t}$ instead of $\io{h_t}$.

We observe that $\io{o_t} \odot \eta(\io{c_t})$, that is output $\io{h_t}$, is a gated $\eta$-view of internal memory $\io{c_t}$. It is used to compute the gates (Eq.~\eqref{eq:lstm.0}) that control the memory update (Eq.~\eqref{eq:lstm.1}), and to update itself (Eq.~\eqref{eq:lstm.2}). Gate $\io{o_t}$ has therefore a critical role in LSTM based sequence learning and encoding. We use a peephole connection~\cite{gers2000recurrent} and our $ca\text{-}attn$ attention with {\bf memory attention dictionary} $A_{mem} \in R^{M_{act} \times D}$ for fine-grained output gating,
\begin{equation}
 \io{o_t} = \sigma(W_o * [ca\text{-}attn(\io{c_t}, A_{mem}), \io{o_{t-1}} \odot \tanh(\io{c_{t-1}})]) \label{eq:output_pooling}
\end{equation}

\subsubsection{Coupling Attention on Memory with Recurrent Attention on Features}
\label{sec:coupling}
To further enhance the memory tracking, we use $\io{y_t}$, that is the attention filtered frame features from Eq.~\eqref{eq:lsta:attn.0}, to internally adapt the attention dictionary for output gating as follows:
\begin{equation}
    A_{mem} \leftarrow A_{mem} + A_o\epsilon(\io{y_t}).\label{eq:coupling}
\end{equation}
We apply the reduction $\epsilon(\io{y_t})$ followed by a linear transform $A_o \in R^{K \times M_{act}}$ to fit the dictionary size. This makes the attention dictionary for fine-grained output gating data dependent.

\subsubsection{LSTA at a Glance}

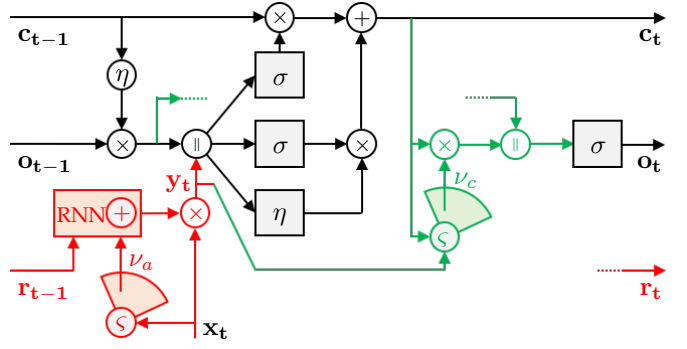
\begin{figure}[!t]
	\centering\includegraphics[width=\linewidth]{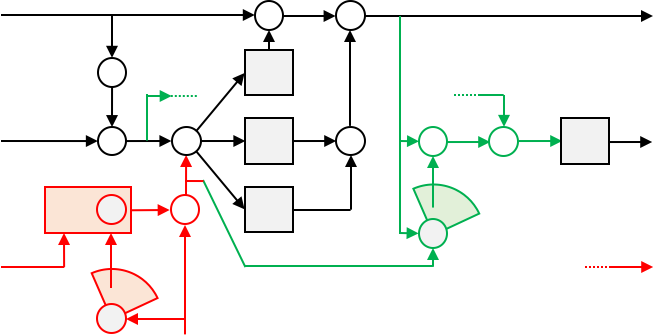}
	\put(-250,113){$\io{c_{t-1}}$}
	\put(-15,113){$\io{c_t}$}
	\put(-250,65){$\io{o_{t-1}}$}
	\put(-15.7,65){$\io{o_t}$}
	\put(-14.7,17){\myr$\io{r_t}$}
	\put(-180,2){$\io{x_t}$} 
	\put(-250,17){\myr$\io{r_{t-1}}$}
	\put(-213,98){$\eta$}
	\put(-213.5,71.7){\footnotesize$\times$}
	\put(-154,120){\footnotesize$\times$}
	\put(-123,119.7){\footnotesize$+$}
	\put(-122.8,71.7){\footnotesize$\times$}
	\put(-153.5,96.7){$\sigma$}
	\put(-153.5,70.7){$\sigma$}
	\put(-153.5,45.6){$\eta$}
	\put(-91.7,71.6){\myg\footnotesize$\times$}
	\put(-33.3,70.8){$\sigma$}
	\put(-213.9,45.8){\myr\footnotesize$+$}
	\put(-185.8,45.7){\myr\footnotesize$\times$}
	\put(-235.2,44.8){\myr\footnotesize RNN}
	\put(-194,58){\myr$\io{y_t}$}
	\put(-208,28){\myr$\nu_a$}
	\put(-212.9,4.7){\myr$\varsigma$}
	\put(-85,59){\myg$\nu_c$}
	\put(-91.2,36.4){\myg$\varsigma$}
	\put(-184,72.6){\tiny$\Vert$}
	\put(-63.5,72.6){\myg\tiny$\Vert$}
	\caption{\ac{lsta} extends \ac{lstm} (black part) with two novel components: recurrent attention and fine-grained output gating. The first (red part, $rca\text{-}attn$ in Eq.~\eqref{eq:lsta:attn.0}) tracks a weight map to focus on relevant feature regions, while the second (green part, Eq.~\eqref{eq:output_pooling}) introduces a high-capacity output gate. At the core of both is a spatial self-attention $\varsigma(\cdot,A)$ that pools parameters from attention dictionary $A$. 
	}
	\label{fig:lsta}
\end{figure}

We now list LSTA equations in detail to ensure reproducibility of our implementation in~\cite{lsta} that can be accessed at \url{https://github.com/swathikirans/LSTA}.
\begin{eqnarray}
\nu_a &\!\!\!\!=\!\!\!\!& \varsigma(\io{x_t}, A_{act})\label{eq:lsta.0}\\
\io{r_t} &\!\!\!\!=\!\!\!\!& convrnn(\nu_a, \io{r_{t-1}})\\
\io{y_t} &\!\!\!\!=\!\!\!\!& softmax(\nu_a + \io{r_t}) \odot \io{x_t}\label{eq:lsta.3}\\
(i_c,f_c,c) &\!\!\!\!=\!\!\!\!& (\sigma,\sigma,\tanh)(W_c \cnv [\io{y_t}, \io{o_{t-1}} \odot \tanh(\io{c_{t-1}})])\label{eq:lsta.4}\\
\io{c_t} &\!\!\!\!=\!\!\!\!& f_c \odot \io{c_{t-1}} + i_c \odot c\label{eq:lsta.5}\\
\nu_c &\!\!\!\!=\!\!\!\!& \varsigma(\io{c_t}, A_{mem} + A_o\epsilon(\io{y_t}))\label{eq:lsta.6}\\
\io{o_t} &\!\!\!\!=\!\!\!\!& \sigma(W_o * [\nu_c\odot\io{c_t}, \io{o_{t-1}} \odot \tanh(\io{c_{t-1}})])\label{eq:lsta.7}
\end{eqnarray}

Fig.~\ref{fig:lsta} illustrates the LSTA. Eqs.~\eqref{eq:lsta.0}-\eqref{eq:lsta.3} implement our recurrent class activation attention $rca\text{-}attn$ from Sec.~\ref{sssec:actenc} (Eqs.~\eqref{eq:lsta.attn.0}-\eqref{eq:lsta.attn.1}; red part in Fig.~\ref{fig:lsta}). Eqs.~\eqref{eq:lsta.6}-\eqref{eq:lsta.7} is our modified output gating from Sec.~\ref{sec:output_pooling} (Eq.~\eqref{eq:output_pooling} with $A_{mem}$ from \eqref{eq:coupling}; green part in Fig.~\ref{fig:lsta}). Bold symbols represent recurrent variables. $\varsigma$ indicates the softmax term of Eq.~\eqref{eq:attn}, that is, $\varsigma(\io{x},A)$ is $ca\text{-}attn(\io{x}, A)$ without $\odot\,\io{x}$.

\subsection{Multi-head Prediction}
\label{ssec:multi}

\begin{figure}
    \centering
    \includegraphics[width=.9\columnwidth]{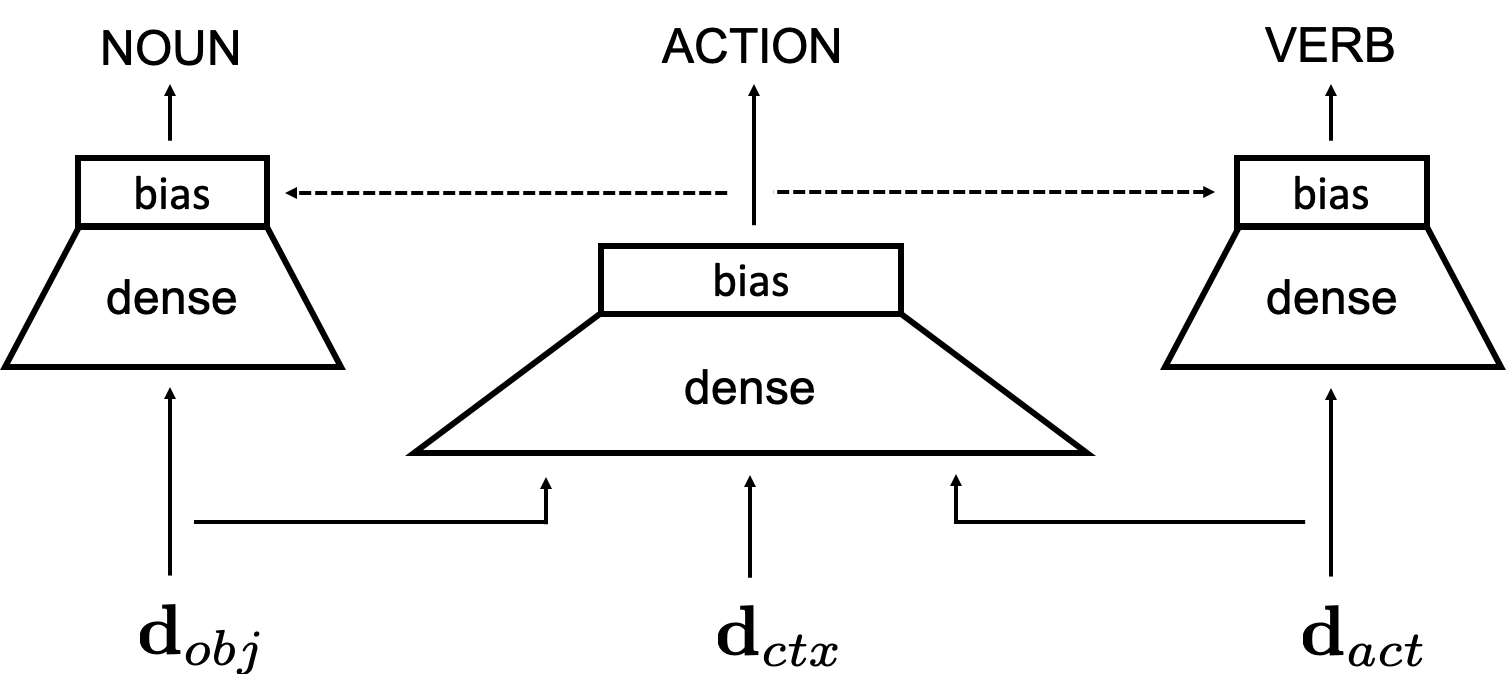}
    \caption{Multi-head prediction in EgoACO. To account for relationships between the tasks, we use the action logits to control the bias of verb and noun prediction, through linear maps (dashed arrows).}
    \label{fig:multihead}
\end{figure}

As discussed earlier, the annotations of egocentric datasets containing object manipulation actions are of the form verb, noun, action labels for each video clip. To leverage such label structure for action recognition, we follow a multi-head prediction strategy to generate the verb, noun, action predictions for each video. To explicitly model the inter-dependencies among the three prediction tasks, we use the prediction scores (logits) of one task to adapt the bias of the classifier of other tasks. This way, a descriptor used to solve one task can influence the learning and prediction of others, \eg, in our implementation $\io{d}_{obj}$ informs about the verb category via the action logits. This is shown in Fig.~\ref{fig:multihead}.

\subsubsection{Noun Prediction}
\label{sec:noun_recognition}
Noun recognition requires locating the active object present in the scene. The active object descriptor ($\io{d}_{obj}$) in Sec.~\ref{ssec:objctx} is trained in a weakly supervised way to locate and attend to the active object present in the video frames.

\subsubsection{Verb Prediction}
\label{sec:verb_recognition}

Identifying the verb action applied on an object requires strong temporal reasoning and the ability to track the relevant spatio-temporal patterns present in the video. For verb recognition, we apply the sum-pooled internal memory state of \ac{lsta}, $\io{d}_{act}$, as the input to a linear classifier.

\subsubsection{Action Prediction}
\label{sec:action_recognition}

In contrast to the majority of the existing works that predict only the verb and noun labels and pair the predictions into an action label, we train our model to predict the action labels as well. Since an action is constituted by the combination of a verb and noun, using $\io{d}_{act}$ and $\io{d}_{obj}$ is a reasonable choice for action prediction. It should be noted that the descriptor $\io{d}_{obj}$ is predicting the noun labels with its own linear weights $B$. In order to predict the action from object-centric descriptor, we use a separate set of linear weights $W \in R^{K \times M_{act}}$. Sharing the attention dictionary, $A_{obj}$, this way allows the network to localize discriminative features of active objects for action prediction. The descriptor $\io{d}_{act}$ is applied to another set of linear layer for generating action predictions. As discussed in Sec.~\ref{ssec:objctx}, the scene context could play a vital role in narrowing down the possible action category candidates. The scene context descriptor, $\io{d}_{ctx}$, maps the input features to the action label space. To obtain the final action prediction for an input video, we average the predictions made by all three descriptors. To model the inter-dependencies of verb, noun and action prediction tasks via bias control as shown in Fig.~\ref{fig:multihead}, we map the action logits to the number of verb and noun classes using linear layers and add them to the verb and noun prediction scores.

%% file: sub/4_results.tex
In this section we describe the experiments and discuss our analysis done for evaluating the performance of the proposed EgoACO framework.

\subsection{Datasets}

We use two large scale egocentric datasets, EPIC-KITCHENS~\cite{epic} and EGTEA Gaze+~\cite{li2018eye}, for benchmarking the proposed approach. Both datasets contain actions carried out in a kitchen environment. 

\noindent{\bf EPIC-KITCHENS} consists of $\sim$39K video clips and $\sim$11.5M frames. The videos are captured in several natural kitchen environments by multiple users, resulting in large variations in appearance of objects, environment and unscripted action patterns executed by the users. The annotations are provided in the form of verb and noun labels, and consist of 125 verbs and 352 noun categories. We combine the verb and noun labels to obtain the action label. There are $2513$ valid action categories in the training set, out of a possible set of $44$K combinations. We report the performance on the two standard test splits, seen (S1) and unseen (S2). S1 consists of video clips from the same environments that are present in the training set. S2 contains videos from sequences of 4 participants that are held out for testing.

\begin{table*}[t]
	\caption{Ablation analysis on the validation split of EPIC-KITCHENS dataset to study the effectiveness of recurrent attention and fine-grained output gating.}
	\centering
	\begin{tabular}{|l|c|c|c|}
		\hline 
		Method & Verb & Noun & Action \\ \hline \hline
		Baseline (ConvLSTM) & 35.16/74.7 & 16/36.57 & 9.87/21.93 \\ \hline
		Baseline + rca-attn & 39.14/73.89 & 16.95/38.19 & 12.25/25.62 \\ \hline
		Baseline + fine-grained output gating & 46/76.94 & 21.32/41.73 & 13.75/28.71 \\ \hline
		Baseline + rca-attn + fine-grained output gating & 45.81/77.47 & 22.36/45.16 & 14.92/30.43 \\ \hline
		LSTA & 47.21/78.38 & 22.19/45.65 & 15.09/30.79 \\ \hline
	\end{tabular}
	\label{tab:lsta_ablation}
\end{table*}

\begin{table}[h]
	\centering 
	\caption{Sensitivity analysis done on the validation split of EPIC-KITCHENS dataset to study the effect of hyper-parameter $D$, the size of the memory attention dictionary $A_{mem}$ used in LSTA for output pooling, Eq.~(21).}
	\begin{tabular}{|c|c|c|c|}
		\hline
		Size of $A_{mem}$  & \multirow{2}{*}{Verb} & \multirow{2}{*}{Noun} & \multirow{2}{*}{Action} \\ 
		in Eq.~(21) &  &  &  \\  \hline\hline 
		100 & 45.24/75.69 & 20.87/42.47 & 13.84/27.8 \\ \hline
		200 & 44.23/76.03 & 20.85/42.78 & 13.56/28.57 \\ \hline
		300 & 47.21/78.38 & 22.19/45.65 & 15.09/30.79 \\ \hline
		400 & 45.2/76.31 & 20.68/41.01 & 14.07/27.02 \\ \hline
		500 & 45.29/76.22 & 20.87/42.57 & 14.28/28.35 \\ \hline
	\end{tabular}
	\label{tab:outputPooling_ablation}
\end{table}

\noindent{\bf EGTEA Gaze+} is the largest egocentric dataset with gaze annotations. Compared to EPIC-KITCHENS, the videos in EGTEA Gaze+ are captured in a single, controlled kitchen environment by 32 people. Thus the appearance variations are less in EGTEA Gaze+ compared to EPIC-KITCHENS. The dataset contains $\sim$10K video clips with verb, noun and action annotations. There are 19 verb classes, 53 noun classes and 106 action classes present in the dataset. The dataset developers provide three standard train/test splits. We evaluate our approach on all three splits in terms of micro (accuracy) and macro (mean class accuracy) metrics. 

\subsection{Implementation Details}
\label{sec:implementation}

\noindent{\bf Backbones.} We use 2D ResNet based CNN (hereafter referred to as R2D) as the backbone in all the experiments. For experiments on the large scale EPIC-KITCHENS dataset we also provide results with R(2+1)D backbone on frame snippets.  For this we change the temporal stride in layers \texttt{conv4\_x} and \texttt{conv5\_x} to $1$ and perform temporal average pooling at the output to reduce the temporal dimension of snippet features to 1 (same as for R2D features).

\noindent{\bf Shared Features.} Decoding the three descriptors $\io{d}_{act}, \io{d}_{ctx}, \io{d}_{obj}$ from a same feature representation $\io{x}$ may be suboptimal. 
Using three separate \ac{cnn}s is computationally expensive and can lead to overfitting. 
In our approach, we share the layers of the ResNet backbone until \verb|conv4_x| and separate \verb|conv5_x| layers for the three different heads. That is in Fig.~\ref{fig:overview}, $\io{x}$ is the output of the shared \verb|conv4_x| and \verb|conv5_x| blocks enter the three descriptor decoding branches. This allows the model to capture different high-level information best suited for the three-fold verb, noun, action classification task, with reduced compute complexity. 

\noindent{\bf Parameters.} $M_{obj}$ and $M_{ctx}$ are selected as the number of noun and action classes for each dataset. The number of hidden units in LSTA, $M_{act}$, is chosen as 512. The attention dictionaries will be initialized with the dense classifier weights of the pre-trained backbone \ac{cnn}, hence, $A_{obj}, A_{act}, A_{ctx}$ with R2D models have cardinality $C=1000$ (from ImageNet) while with R(2+1)D model we have $C=400$ (from IG-Kinetics-65M). The size of $A_{mem}$, the memory attention dictionary  of \ac{lsta}, is set to $D=300$.

\noindent{\bf Training.} R2D \acp{cnn} were pre-trained on ImageNet while R(2+1)D was pre-trained on IG-Kinetics-65M~\cite{weakly_supervised}. Attention dictionaries $A_{obj}, A_{act}, A_{ctx}$ are initialized with the dense classifier weights of the pre-trained backbone \ac{cnn}. All other parameters are initialized using Kaiming initialization~\cite{kaiming_init}. EgoACO is trained end-to-end with the loss
\begin{equation}
L = L_N + L_V + L_A \label{multitask_loss}
\end{equation}
where $L_N, L_V, L_A$ represent the categorical cross-entropy loss of noun, verb, action predictions. We train the network in three stages. In the first stage, the weights of the backbone \ac{cnn} are frozen while all the other layers are updated. In stages 2 and 3, the parameters in \verb|conv5_x| and \verb|conv4_x| of the backbone are trained together with the layers trained in stage 1. Such multi-stage training strategy stabilizes the training process and has been applied in previous works~\cite{girdhar2017actionvlad, tdn, wang2018videos, sap}. \ac{sgd} with momentum $0.9$ and weight decay $5\times10^{-4}$ is used as the optimizer. We use a cosine learning rate schedule~\cite{loshchilov2016sgdr} without warm restarts for adjusting the learning rate during training. Stages 1 and 2 are trained with an initial learning rate of $10^{-2}$ and for 60 epochs while stage 3 is trained for 30 epochs with an initial learning rate of $10^{-4}$. A dropout of $0.5$ is used to prevent overfitting. A batch size of $32$ is used in all three training stages. We follow the sparse sampling strategy of TSN in~\cite{tsn} for model training and testing. A video clip is first divided into 20 equal-length segments. A frame is then randomly sampled from each segment. The 20 sampled frames are applied as input to R2D based models while for R(2+1)D based models we use the three consecutive frames around each sampled frame to form the 20 input snippets. For spatial augmentation, we use corner cropping and scale jittering. We first scale the frame such that the size of the shorter side is $256$ for R2D models and $128$ for R(2+1) models. Then we randomly crop from either the corners or the center. The dimension of the cropped clip is randomly selected from \{$256$, $224$, $192$, $168$\} for R2D and from \{$128$, $112$, $96$, $84$\} for R(2+1)D models. The cropped region is then rescaled to $224$ for R2D models and $112$ for R(2+1)D based models. The cropped frames are further randomly flipped horizontally. 

\noindent{\bf Testing.} From each clip we sample 20 frames/snippets twice, to form two input sequences. The first sequence is made of the central frame/snippet of each segment while for the second sequence we use the first frame/snippet of each segment. 
The selected frames are first scaled such that the shorter size is $256$ for R2D models and $128$ for R(2+1)D models. We then crop from the center of the frame with a dimension of $224\times224$ for R2D based models while for R(2+1)D based models, the cropped frame size is $112\times112$.  The scores obtained from the two input sequences are averaged to obtain the output prediction scores.

We follow the same settings for both datasets used in this study. Source code of our implementation can be found online at \url{https://github.com/swathikirans/EgoACO}.

\subsection{Model Analysis}
\label{sec:ablation}

The model analysis is done on the validation split, proposed in~\cite{baradel2018object}, of EPIC-KITCHENS dataset. We use 2D ResNet with depth 34 (R2D-34) as the \ac{cnn} backbone for all the ablation studies. In Sec.~\ref{sec:lsta_ablation} we analyze the impact of recurrent attention (Sec.~\ref{sssec:actenc}) and fine-grained output gating (Sec.~\ref{sec:output_pooling}) on ConvLSTM. We study the effectiveness of object and scene context pooling (Sec.~\ref{ssec:objctx}) along with an analysis of the effectiveness of our design choices in modeling the inter-dependencies among the three prediction tasks in Sec.~\ref{sec:egoaco_ablation}. In Sec.~\ref{sec:multitask_ablation} we evaluate the effectiveness of three-fold supervision via multi-task learning in training EgoACO. Sec.~\ref{sec:attentional_pooling_ablation} presents the comparison between our dictionary-based self attention mechanism against attentional pooling~\cite{girdhar2017attentional} and class activation mapping~\cite{zhou15cnnlocalization}. We also visualize attention maps generated by EgoACO in Sec.~\ref{sec:vis_ablation}. In all the tables reporting the results of our model analysis, both Top-1 and Top-5 recognition accuracies are shown (Top-1/Top-5).

\subsubsection{Ablation on LSTA}
\label{sec:lsta_ablation}
Tab.~\ref{tab:lsta_ablation} shows the results of the study done on \ac{lsta}. We choose a model with ConvLSTM as the baseline. We predict verb, noun, action labels from the internal memory state of ConvLSTM. Next we added recurrent attention (rca-attn, Sec.~\ref{sssec:actenc}) to the ConvLSTM baseline. We observe  improvements in the performance of all three tasks, over the baseline. The most improvement was observed for verb and action predictions. Detailed analysis shows that the attention mechanism enables the network to correctly classify actions with objects that are similar in appearance such as \verb|take_plate| and \verb|take_bowl|, \verb|put_lid| and \verb|put-down_pan|.

Replacing the standard output gating of ConvLSTM (Eq.~\ref{eq:lstm.2}) with our fine-grained output gating (Eq.~\ref{eq:output_pooling}) described in Sec.~\ref{sec:output_pooling} results in an increase of $+10.84\%$, $+5.32\%$ and $+3.88\%$ for verb, noun and action recognition accuracies, respectively. Analysing the most improved classes by adding our enhanced output gating, we observe that the network improved its ability to recognize verbs that are similar in execution such as \verb|peel| and \verb|remove|, \verb|put| and \verb|pour|, \verb|put| and \verb|close|.

Then, we added both rca-attn and fine-grained output gating to the baseline and observed an improvement of $+5.05\%$ in action prediction. Analysis of the improved classes over the baseline shows that it is able to distinguish between verbs and nouns that are similar in action and appearance, respectively, such as \verb|put| and \verb|pour|, \verb|scrape| and \verb|wash|, \verb|food| and \verb|pasta|, \verb|pan| and \verb|pot|. 

We finally coupled the attention on memory with recurrent attention on features as described in Sec.~\ref{sec:coupling}, resulting in \ac{lsta}. This resulted in a performance gain over baseline of $+12.05\%$, $+6.19\%$ and $+5.22\%$ for verb, noun and action recognition accuracies, respectively. Analysis of the most improved classes reveals that \ac{lsta} is able to discriminate actions involving similar objects such as \verb|peel_potato| and \verb|put-down_potato|, \verb|close_lid| and \verb|put_lid|, as well as actions involving similar \verb|verb| patterns such as \verb|wash_knife| and \verb|wash_spoon|, \verb|wash_plate| and \verb|wash_bowl|.

One notable feature of LSTA is the improved memory propagation using attention mechanism, as described in Sec.~\ref{sec:output_pooling}. Tab.~\ref{tab:outputPooling_ablation} reports the result of sensitivity analysis conducted on LSTA by varying the size of the memory attention dictionary, $D$. As can be seen from the table, the recognition performance increases as the dictionary size is increased from 100 to 300 followed by a gradual reduction. Following this result, we choose $D=300$ in our experiments.

\begin{table}[h]
	\caption{Ablation analysis on the validation split of EPIC-KITCHENS dataset to study the effectiveness of active object and scene context pooling.}
	\centering
	\begin{tabular}{|l|c|c|c|}
		\hline
		Method & Verb & Noun & Action \\ \hline \hline
		\ac{lsta} & 47.21/78.38 & 22.19/45.65 & 15.09/30.79 \\ \hline
		\ac{lsta} + active object & 47.43/77.39 & 22.67/45.65 & 15.55/32.23 \\ \hline
		\ac{lsta} + scene context & 46.7/77.66 & 22.34/44.12 & 15.3/32.30 \\ \hline
		EgoACO & 48.25/79.09 & 22.78/45.71 & 16.63/33.31 \\ \hline
	\end{tabular}
	\label{tab:model_ablation}
\end{table}

\subsubsection{Ablation on EgoACO}
\label{sec:egoaco_ablation}
Tab.~\ref{tab:model_ablation} reports the results of the ablation study done for analyzing the impact of active object and scene context pooling described in Sec.~\ref{ssec:objctx}. A \ac{cnn}-\ac{lsta} model is chosen as the baseline for this study. We first added the object-centric stream to the baseline, resulting in $+0.22\%$, $+0.48\%$ and $+0.46\%$ for verb, noun and action prediction performances, respectively. Then we applied the scene context stream to the baseline. This results in an improvement of $+0.21\%$ for action recognition accuracy. Adding both object-centric and scene context branches to the baseline results in our final model, which provides a performance gain of $+1.04\%$, $+0.59\%$ and $+1.54\%$ for verb, noun and action, respectively. 

We report further ablation study to validate the importance of object features stream and scene context stream. In this, we first remove the object stream and use scene context stream for noun prediction. \ie, noun is predicted from $\io{d_{ctx}}$ instead of $\io{d}_{obj}$. This results in a drop of $-3.37\%$ ($13.26\%$) in action recognition accuracy. By replacing $\io{d}_{ctx}$ with $\io{d}_{obj}$, the action recognition accuracy drops by an absolute $-2.52\%$, from $16.63\%$ to $14.11\%$. This shows that the two descriptors encode different information from the video, due to the variations in attention computation and how the descriptors across frames are pooled, as explained in Sec.~\ref{ssec:objctx}.

Next we evaluated the effectiveness of bias control, as introduced in Sec.~\ref{ssec:multi}, in modeling the inter-dependencies among the three prediction tasks. In this, we remove the application of linearly mapped action logits from the verb and noun classifiers. This results in an accuracy of $43.76\%$, $10.22\%$ and $13.07\%$ for verb, noun and action recognition, respectively. The larger drop in noun accuracy ($-12.56\%$) shows that allowing the action score (generated from $\io{d}_{act}$ which receives direct supervision from verb and action labels) to influence the noun prediction allows EgoACO to better model the dependencies present in the label structure.  This is more relevant in the case of egocentric videos, where the scene is cluttered with many objects, of which only one is the active object. This bias control allows the model to discard those objects present in the scene that cannot co-occur with the verb to form an action category, thereby improving noun prediction performance.

\begin{table}[h]
	\centering 
	\caption{Analysis on the validation split of EPIC-KITCHENS dataset to validate the effectiveness of training EgoACO in a multi-task fashion.}
	\begin{tabular}{|c|c|c|c|}
		\hline
		Supervision & Verb & Noun & Action \\ \hline \hline
		V + N & 44.31/74.53 & 19.09/38.46 & 12.27/32.46 \\ \hline
		A & 38.17/68.23 & 16.15/37.44 & 11.65/24.26 \\ \hline
		V + N + A & 48.25/79.09 & 22.78/45.71 & 16.63/33.31 \\ \hline
	\end{tabular}
	\label{tab:multitask_ablation}
\end{table}

\subsubsection{Analysis on the Effect of Multi-task Learning}
\label{sec:multitask_ablation}	
As mentioned before, actions in egocentric datasets are generally defined as a combination of verb-noun pairs. Thus, one can train a network to predict the action label alone or verb and noun labels. In this section, we study how one can effectively utilize the different annotations available in the dataset. Tab.~\ref{tab:multitask_ablation} reports the result of this study. The results are reported by changing the supervision provided to the network without altering its architecture. From the table, one can see that supervision using all three annotations results in a significant improvement in action recognition accuracy. Using all the three annotations allows the network to model the inter-dependencies among the labels and to capture the semantics from the shared pool of features, thereby improving its recognition performance.

\subsubsection{Comparison to Related Approaches}
\label{sec:attentional_pooling_ablation}

In this section, we compare class activation pooling with two related approaches, attentional pooling~\cite{girdhar2017attentional} and class activation mapping (CAM)~\cite{zhou15cnnlocalization}, for attention generation.

By changing the cardinality of the attention dictionary weights $A$ in Eq.~\ref{eq:cap.0}, class activation pooling collapses to attentional pooling~\cite{girdhar2017attentional}. Precisely, attentional pooling is CAP with a dictionary of one element, $A = \{\io{a} \in R^{K\times 1}\}$ and thus $\io{a}_{c^*}=\io{a}$ in Eq.~\ref{eq:cap.0}. A single free parameter \io{a} is responsible for attending to the discriminant features in the scene. To study the effectiveness of our dictionary learning approach, we replace class activation pooling with attentional pooling for generating active object and scene context descriptors. We also update the recurrent attention and memory attention mechanisms in \ac{lsta} with attentional pooling.  A drop of $-2.77$\% (13.86\%) in action recognition accuracy, compared to EgoACO, is observed. This shows the effectiveness of our flexible model consisting of a dictionary of weights.

CAM~\cite{zhou15cnnlocalization} is a special case of CAP where the classifier weights used to generate prediction scores serve as attention weights too. Precisely, $B=A$ in Eq.~(\ref{eq:cap.0}). We replace CAP in EgoACO with CAM, that is, we use the noun classifier weights as the active object dictionary and action classifier weights as the scene context dictionary. We further update \ac{lsta} to use the verb classifier weights as the dictionary for recurrent attention. Using CAM for attention resulted in a degradation of $-3.56\%$ ($13.07\%$) in action recognition accuracy compared to CAP ($16.63\%$). This validates the effectiveness of CAP over CAM as a self-attention mechanism.

\begin{figure*}[t]
    \centering
    \begin{subfigure}[b]{\textwidth}
    \centering
    \raisebox{0.22in}{\rotatebox[origin=t]{90}{{\small Object}}} \hskip -0.54mm
    \includegraphics[scale=0.21]{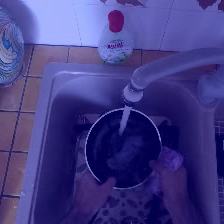}
    \includegraphics[scale=0.21]{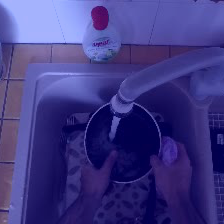}
    \includegraphics[scale=0.21]{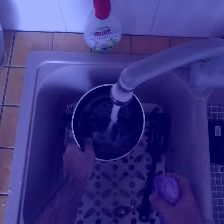}
    \includegraphics[scale=0.21]{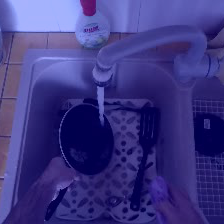}
    \includegraphics[scale=0.21]{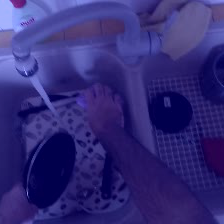}
    \includegraphics[scale=0.21]{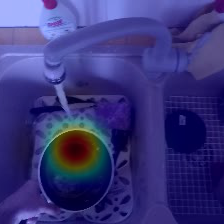}
    \includegraphics[scale=0.21]{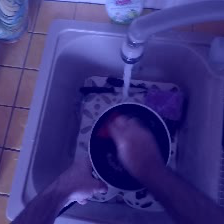}
    \includegraphics[scale=0.21]{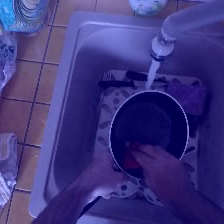}
    \includegraphics[scale=0.21]{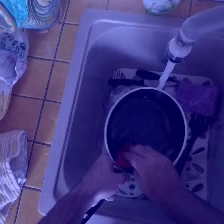}
    \includegraphics[scale=0.21]{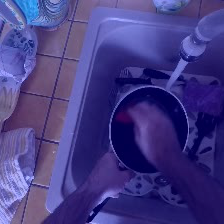} \\ 
    \vskip 1mm
    \raisebox{0.22in}{\rotatebox[origin=t]{90}{{\small Context}}}
    \includegraphics[scale=0.21]{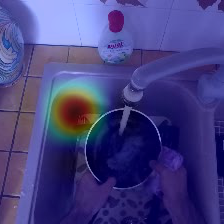}
    \includegraphics[scale=0.21]{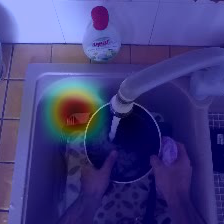}
    \includegraphics[scale=0.21]{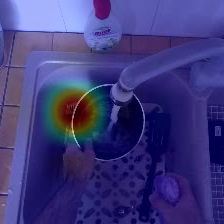}
    \includegraphics[scale=0.21]{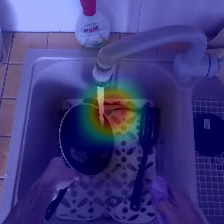}
    \includegraphics[scale=0.21]{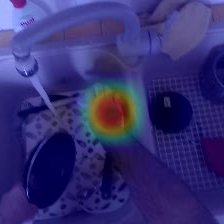}
    \includegraphics[scale=0.21]{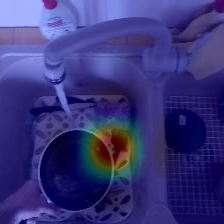}
    \includegraphics[scale=0.21]{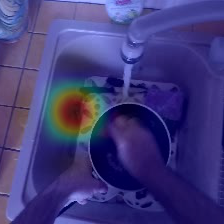}
    \includegraphics[scale=0.21]{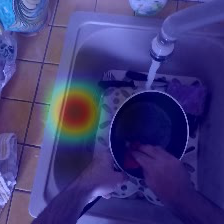}
    \includegraphics[scale=0.21]{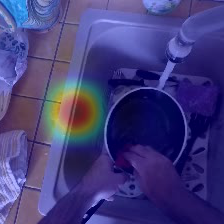}
    \includegraphics[scale=0.21]{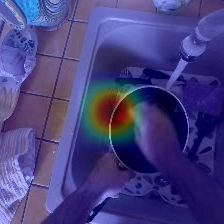} \\ 
    \vskip 1mm
    \raisebox{0.22in}{\rotatebox[origin=t]{90}{{\small Action}}}
    \includegraphics[scale=0.21]{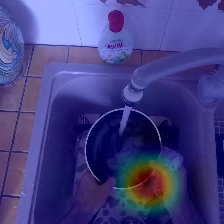}
    \includegraphics[scale=0.21]{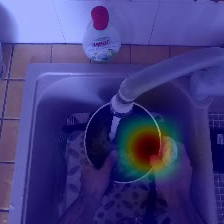}
    \includegraphics[scale=0.21]{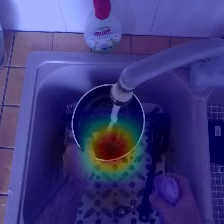}
    \includegraphics[scale=0.21]{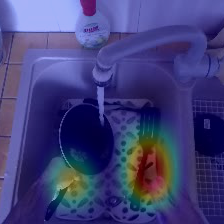}
    \includegraphics[scale=0.21]{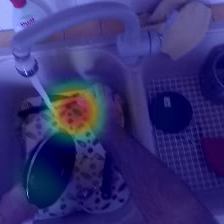}
    \includegraphics[scale=0.21]{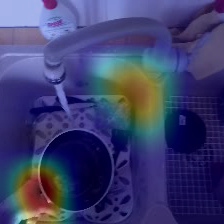}
    \includegraphics[scale=0.21]{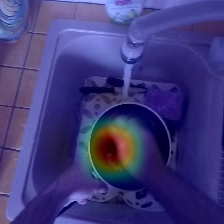}
    \includegraphics[scale=0.21]{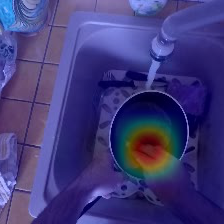}
    \includegraphics[scale=0.21]{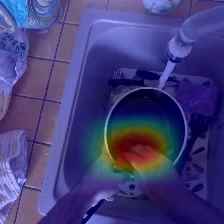}
    \includegraphics[scale=0.21]{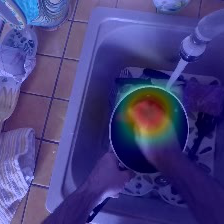}
    \vskip -1mm
    \caption{\texttt{wash\_pan}}
    \label{fig:wash_pan}
    \end{subfigure}
    \vskip 2mm
    \begin{subfigure}[b]{\textwidth}
    \centering
    \raisebox{0.22in}{\rotatebox[origin=t]{90}{{\small Object}}} \hskip -0.54mm
    \includegraphics[scale=0.21]{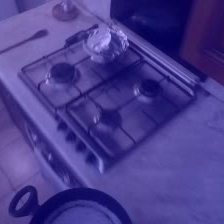}
    \includegraphics[scale=0.21]{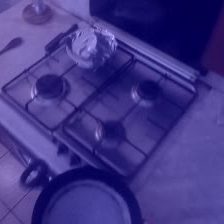}
    \includegraphics[scale=0.21]{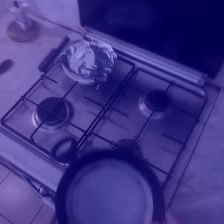}
    \includegraphics[scale=0.21]{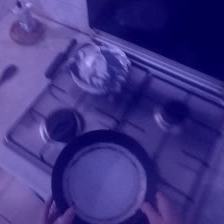}
    \includegraphics[scale=0.21]{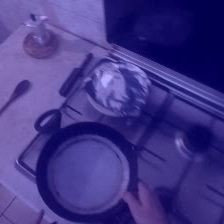}
    \includegraphics[scale=0.21]{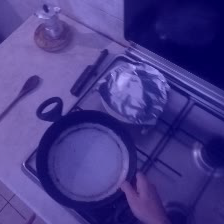}
    \includegraphics[scale=0.21]{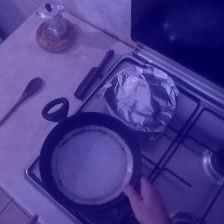}
    \includegraphics[scale=0.21]{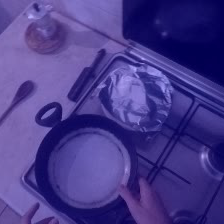}
    \includegraphics[scale=0.21]{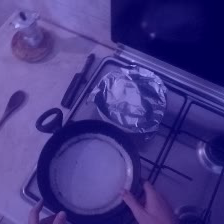}
    \includegraphics[scale=0.21]{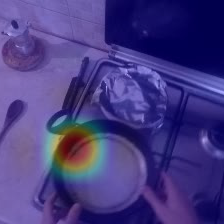} \\ 
    \vskip 1mm
    \raisebox{0.22in}{\rotatebox[origin=t]{90}{{\small Context}}}
    \includegraphics[scale=0.21]{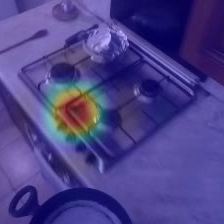}
    \includegraphics[scale=0.21]{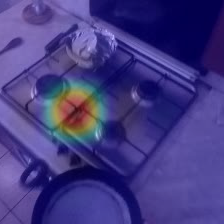}
    \includegraphics[scale=0.21]{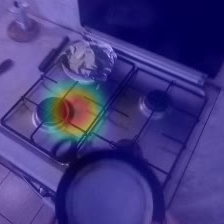}
    \includegraphics[scale=0.21]{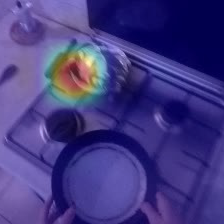}
    \includegraphics[scale=0.21]{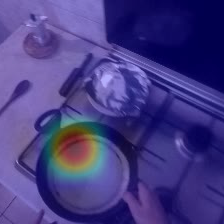}
    \includegraphics[scale=0.21]{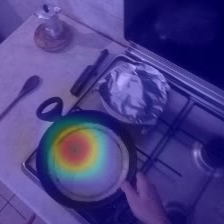}
    \includegraphics[scale=0.21]{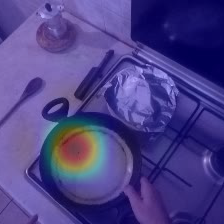}
    \includegraphics[scale=0.21]{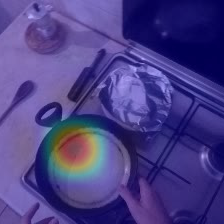}
    \includegraphics[scale=0.21]{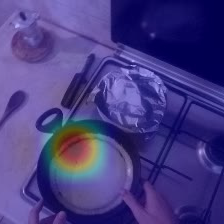}
    \includegraphics[scale=0.21]{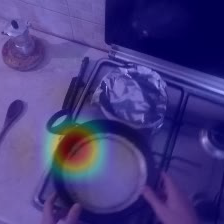} \\ 
    \vskip 1mm
    \raisebox{0.22in}{\rotatebox[origin=t]{90}{{\small Action}}}
    \includegraphics[scale=0.21]{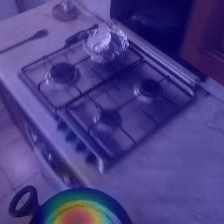}
    \includegraphics[scale=0.21]{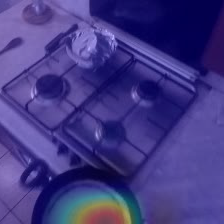}
    \includegraphics[scale=0.21]{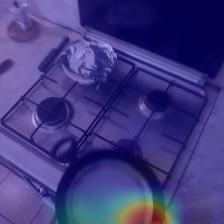}
    \includegraphics[scale=0.21]{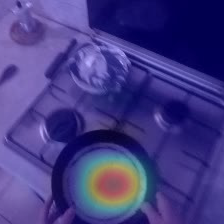}
    \includegraphics[scale=0.21]{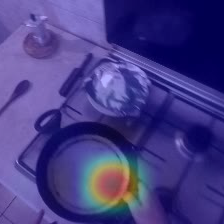}
    \includegraphics[scale=0.21]{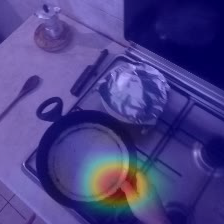}
    \includegraphics[scale=0.21]{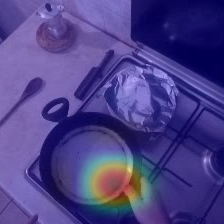}
    \includegraphics[scale=0.21]{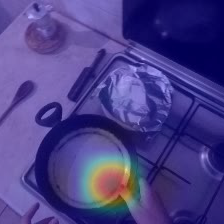}
    \includegraphics[scale=0.21]{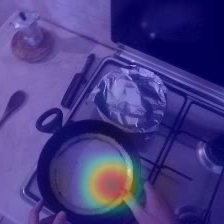}
    \includegraphics[scale=0.21]{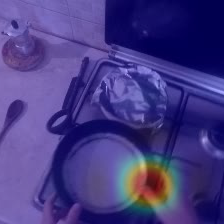}
    \vskip -1mm
    \caption{\texttt{put-down\_pan}}
    \label{fig:put_down_pan}
    \end{subfigure}
    \vskip 2mm
    \begin{subfigure}[b]{\textwidth}
    \centering
    \raisebox{0.22in}{\rotatebox[origin=t]{90}{{\small Object}}} \hskip -0.54mm
    \includegraphics[scale=0.21]{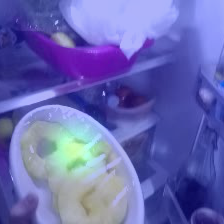}
    \includegraphics[scale=0.21]{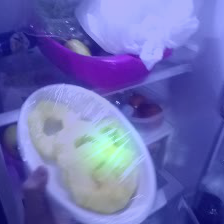}
    \includegraphics[scale=0.21]{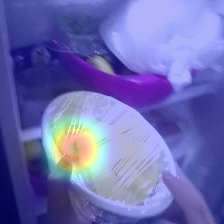}
    \includegraphics[scale=0.21]{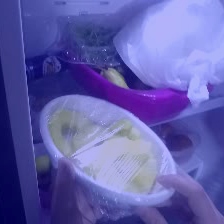}
    \includegraphics[scale=0.21]{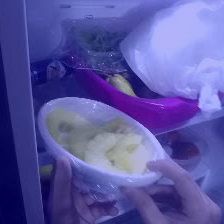}
    \includegraphics[scale=0.21]{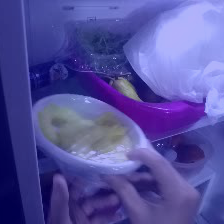}
    \includegraphics[scale=0.21]{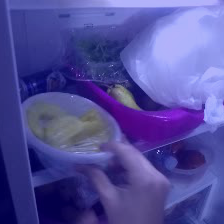}
    \includegraphics[scale=0.21]{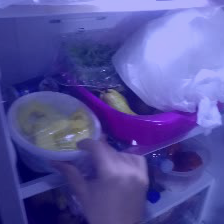}
    \includegraphics[scale=0.21]{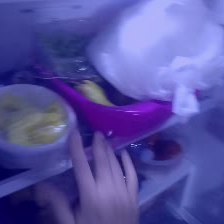}
    \includegraphics[scale=0.21]{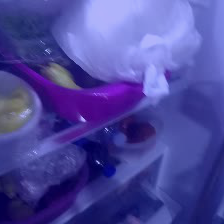} \\ 
    \vskip 1mm
    \raisebox{0.22in}{\rotatebox[origin=t]{90}{{\small Context}}}
    \includegraphics[scale=0.21]{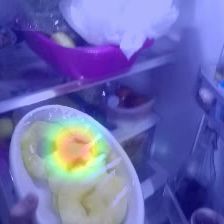}
    \includegraphics[scale=0.21]{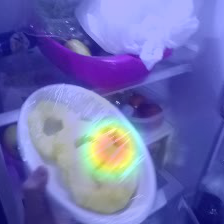}
    \includegraphics[scale=0.21]{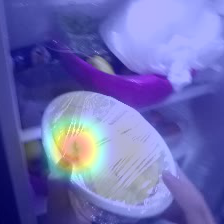}
    \includegraphics[scale=0.21]{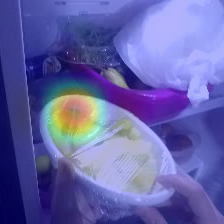}
    \includegraphics[scale=0.21]{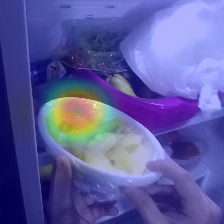}
    \includegraphics[scale=0.21]{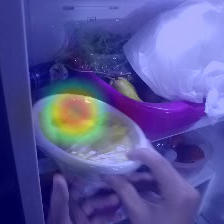}
    \includegraphics[scale=0.21]{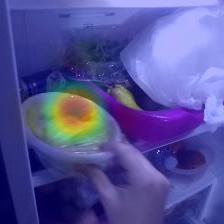}
    \includegraphics[scale=0.21]{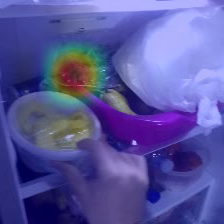}
    \includegraphics[scale=0.21]{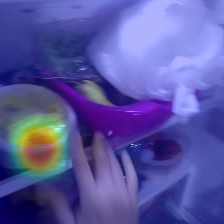}
    \includegraphics[scale=0.21]{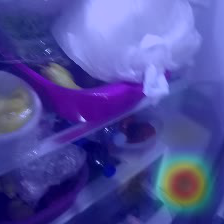} \\ 
    \vskip 1mm
    \raisebox{0.22in}{\rotatebox[origin=t]{90}{{\small Action}}}
    \includegraphics[scale=0.21]{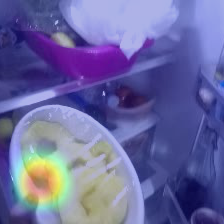}
    \includegraphics[scale=0.21]{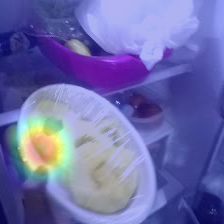}
    \includegraphics[scale=0.21]{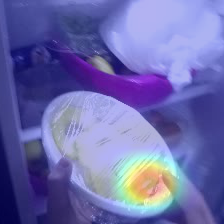}
    \includegraphics[scale=0.21]{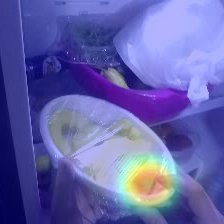}
    \includegraphics[scale=0.21]{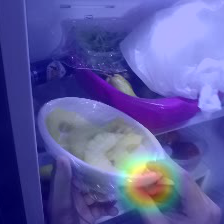}
    \includegraphics[scale=0.21]{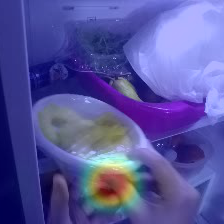}
    \includegraphics[scale=0.21]{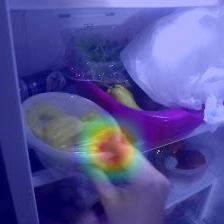}
    \includegraphics[scale=0.21]{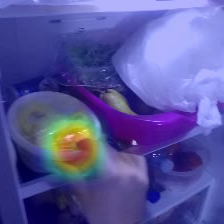}
    \includegraphics[scale=0.21]{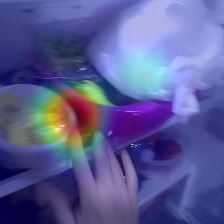}
    \includegraphics[scale=0.21]{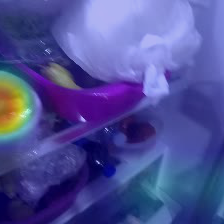}
    \vskip -1mm
    \caption{\texttt{put\_pineapple}}
    \label{fig:put_pineapple}
    \end{subfigure}
    \caption{Attention maps generated by $\io{d}_{obj}$ (first row), $\io{d}_{ctx}$ (second) and LSTA (third) for three samples from the validation split of EPIC-KITCHENS. We show 10 frames uniformly sampled from the 20 frames applied as input to the network.}
	\label{fig:vis_epic}
\end{figure*}

\subsubsection{Attention Maps Visualization}
\label{sec:vis_ablation}

In Fig.~\ref{fig:vis_epic} we visualize the attention maps generated by our model on samples from the validation split of EPIC-KITCHENS dataset. From the figures one can see that the recurrent attention in LSTA tracks hands or hand-object interface, enabling the network to improve verb prediction. On the other hand, the object stream attention is able to localize the discriminative regions that are important for recognizing the noun label, such as regions of the pan. The context stream attention is looking at regions in each frame, independently, that can assist in action prediction such as the hob and pan in Fig.~\ref{fig:put_down_pan} and regions of the sink in Fig.~\ref{fig:wash_pan}. Fig.~\ref{fig:put_pineapple} show the attention maps generated by the network for a sample with wrong predictions. The ground truth labels for verb, noun and action of this sample are \texttt{put}, \texttt{pineapple} and \texttt{put\_pineapple}, respectively. The categories predicted by the network are \texttt{put}, \texttt{bowl} and \texttt{put\_bowl}. In this example, the object stream attention is stronger in the third frame and it is near the hand-object (bowl) rather than on the regions containing pineapple. This forces the network to make a wrong prediction for the object category. It should be noted that the three descriptors $\io{d}_{obj}, \io{d}_{ctx}, \io{d}_{act}$ are computed from different features of the backbone CNN following different spatio-temporal aggregation. Thus their descriptors will be different even if the attention visualizations look similar.

\begin{table*}[t]
	\centering 
	\caption{Comparison of recognition accuracies with state-of-the-art on EPIC-KITCHENS dataset. $^\dagger$: Taken from~\cite{epic_evaluation}; $^\ddagger$: Also uses Faster R-CNN for object detection; $^{\dagger\dagger}$: Taken from~\cite{weakly_supervised}, using a lower resolution ($112\times112$) than that used to report the results ($128\times128$); $^*$: Computed using publicly available implementation.}
	\begin{tabular}{c|c|c|c|c|c|c|c|c|c|c|c|}
		\hline
	&	\multirow{2}{*}{\textbf{Method}} &
	\multirow{2}{*}{\textbf{Backbone}} & \multirow{2}{*}{\textbf{Pre-train}} & \multicolumn{3}{c|}{\textbf{S1}} & \multicolumn{3}{c|}{\textbf{S2}} & \multirow{2}{*}{\textbf{FLOPs (G)}} & \multirow{2}{*}{\textbf{Params (M)}} \\ \cline{5-10}
	&		& & & \textbf{Verb} & \textbf{Noun} & \textbf{Action} & \textbf{Verb} & \textbf{Noun} & \textbf{Action} & & \\ \hline \hline
		\multirow{5}{*}{\rotatebox[origin=c]{90}{Multimodal}} &	TSN~\cite{tsn}$^\dagger$ & R2D-50 & ImNet & 55.5 & 41.28 & 26.89 & 45.75 & 25.13 & 15.40 & 68.29$\times$10$^*$ & 48.93 \\ \cline{2-12}
		&	TSM~\cite{tsm}$^\dagger$ & R2D-50 & ImNet & 62.37 & 41.88 & 29.90 & 51.96 & 25.61 & 17.38 & 68.29$\times$10$^*$ & 48.99 \\ \cline{2-12}
		&	M-TRN~\cite{trn}$^\dagger$ & R2D-50 & ImNet & 62.68 & 39.82 & 29.41 & 52.03 & 25.88 & 17.86 & 68.29$\times$10$^*$ & 52.49 \\ \cline{2-12}
		&	TBN~\cite{tbn} & BNInc & ImNet+UCF & 64.75 & 46.03 & 34.80 & 52.69 & 27.86 & 19.06 & 173.74$^*$ & 32.64 \\ \cline{2-12}
		&	RU-LSTM~\cite{rulstm} & BNInc$^\ddagger$ & ImNet & 56.93 & 43.05 & 33.06 & 43.67 & 26.77 & 19.49 & - & - \\ \hline 
		\multirow{12}{*}{\rotatebox[origin=c]{90}{RGB only}} &	TSN~\cite{tsn}$^\dagger$ & R2D-50 & ImNet & 49.71 & 39.85 & 23.97 & 36.70 & 23.11 & 12.77 & 33.12$\times$10 & 24.48 \\ \cline{2-12}
		&	TSM~\cite{tsm}$^\dagger$ & R2D-50 & ImNet & 57.88 & 40.84 & 28.22 & 43.50 & 23.32 & 14.99 & 33.12$\times$10 & 24.48 \\ \cline{2-12}
		&	M-TRN~\cite{trn}$^\dagger$ & R2D-50 & ImNet & 60.16 & 38.36 & 28.23 & 46.94 & 24.41 & 16.32 & 33.12$\times$10 & 25.33 \\ \cline{2-12}
		&	LSTA~\cite{lsta} & R2D-34 & ImNet & 58.25 & 38.93 & 30.16 & 45.51 & 23.46 & 15.88 & 92.03 & 44.28 \\ \cline{2-12}
		&	LFB~\cite{lfb} & R3D-50-NL$^\ddagger$ & Kin & 60.0 & 45 & 32.7 & 50.9 & 31.5 & 21.2 & - & - \\ \cline{2-12}	
		&	R(2+1)D~\cite{weakly_supervised} &R(2+1)D-34 & IG-Kin-65M & 63.3 & 46.3 & 34.4 & 55.5 & 33.6 & 23.7 & 399.56$\times$10$^*$ & 127.4 \\ \cline{2-12}
		&	R(2+1)D~\cite{weakly_supervised} & R(2+1)D-152 & IG-Kin-65M & 65.2 & 45.1 & 34.5 & 57.3 & 35.7 & 25.6 & $>$504$\times$10$^{\dagger\dagger}$ & 236 \\ \cline{2-12}
		&	SAP~\cite{sap} & R3D-50$^\ddagger$ & Kin & 63.2 & 48.3 & 34.80 & 53.2 & 33.0 & 23.9 & - & - \\ \cline{2-12}\cline{2-12}
		& \multirow{4}{*}{EgoACO} & R2D-34 & ImNet & 60.38 & 40.33 & 31.53 & 47.97 & 25.30 & 18.33 & 117.77$\times$2 & 74.2 \\ \cline{3-12}
		&	 & R2D-50 & ImNet & 61.97 & 41.83 & 32.11 & 47.29 & 26.39 & 18.71 & 154.37$\times$2 & 116.22 \\ \cline{3-12}
		&	 & R2D-101 & ImNet & 61.85 & 42.03 & 33.22 & 51.01 & 28.99 & 21.41 & 228.92$\times$2 & 135.21 \\ \cline{3-12}
		&	 & R(2+1)D-34 & IG-Kin-65M &  66.88 & 46.28 & 37.32 & 55.86 & 31.44 & 23.35 & 643.56$\times$2 & 167.02 \\ \hline
	\end{tabular}
	\label{tab:epic_sota}
\end{table*}

\subsection{Comparison with State-of-the-Art}
\label{sec:comparison}

We compare the performance of EgoACO with state-of-the-art approaches on EPIC-KITCHENS in Tab.~\ref{tab:epic_sota}. The first part of the table lists the methods that use multiple data modalities such as optical flow and/or audio while the second section consists of methods that use RGB frames alone. We report the performance of our method using R2D-34, R2D-50, R2D-101 and R(2+1)D-34 backbone \acp{cnn}. With the same backbone, R2D-34, we improve the performance of \ac{lsta} from our prior work~\cite{lsta} by $+1.37\%$ and $2.45\%$ on action for S1 and S2 splits, respectively. The larger improvement in S2 split shows that our method is more generalizable. Compared to popular action recognition approaches such as TSN~\cite{tsn}, M-TRN~\cite{trn} and TSM~\cite{tsm} and their corresponding two-stream approaches, with the same backbone CNN R2D-50, EgoACO results in a better recognition performance on both S1 and S2 splits with less computational complexity.

With R2D-101 backbone, we improve action recognition performance over all methods that use multiple data modalities on split S2 while on S1, we rank second to TBN~\cite{tbn} which uses audio modality in addition to RGB and optical flow. Considering the methods that use RGB alone, our approach, with R2D-101, is superior to LFB~\cite{lfb}, which uses an object detector and is pre-trained on Kinetics dataset.

With an R(2+1)D-34~\cite{r2+1d} backbone pre-trained on IG-Kinetics-65M dataset, we outperform all state-of-the-art approaches by a large margin on S1 split. SAP~\cite{sap} uses two R3D-50 backbone CNNs for verb and noun feature generation along with a Faster R-CNN~\cite{ren2015faster} based on ResNeXt-101-FPN~\cite{resnext} for object detection. Despite using this high capacity model that operates on a larger number of frames ($64$) from the video, SAP obtains an action recognition accuracy of $34.8\%$, compared to the $37.32\%$ ($+2.52\%$) of EgoACO on S1 split while both approaches perform comparably on S2 split. Compared to R(2+1)D-34~\cite{weakly_supervised} and R(2+1)D-152~\cite{weakly_supervised}, EgoACO results in an improvement of $+2.92\%$ and $+2.82\%$ in action recognition accuracy on S1 split, with far less computational complexity. On S2 split, EgoACO is comparable to R(2+1)D-34 while R(2+1)D-152 outperforms EgoACO by $2.25\%$. It should be noted that the R(2+1)D approach from~\cite{weakly_supervised} independently trains two different backbone CNNs for verb and noun predictions. On the other hand, EgoACO relies on the shared features extracted from a single backbone CNN. The improvement obtained by plugging in EgoACO head on top of the R(2+1)D-34 backbone shows that EgoACO is capable of leveraging the inter-dependencies among the three different annotations. This is made possible by the self-attention mechanism based on CAP that enables the network to focus on different regions in the video for generating effective video descriptors.

In Tab.~\ref{tab:egtea_sota}, we report the performance of our method on EGTEA Gaze+ dataset and compare against state-of-the-art approaches. We report both micro and macro metrics on all three splits of the dataset. We also report updated results (micro metrics) of our previous approaches, Ego-RNN~\cite{egornn} and \ac{lsta}~\cite{lsta}. Methods that use multiple data modalities and those using RGB frames are separated into two sections. We report the results using R2D-101 \ac{cnn}. The network is fine-tuned from the EPIC-KITCHENS pre-trained model. From the table, it can be seen that our method outperforms all other approaches, including those that use multiple modalities of data~\cite{simonyan2014two, tsn, carreira2017quo, egornn, rulstm,li2018eye, huang2019mutual, lu2019learning} and additional annotations in the form of gaze~\cite{li2018eye, huang2019mutual, lu2019learning} and object detections~\cite{sap, rulstm}.

\begin{table*}[t]
	\centering
	\caption{Comparison of recognition accuracies with state-of-the-art on EGTEA Gaze+ dataset. $^\dagger$: Taken from~\cite{li2018eye}; $^{\dagger\dagger}$: Implemented by us; $^\ddagger$: Also uses Faster R-CNN for object detection.}
	\begin{tabular}{|c|c|c|c|c|c|c|c|c|c|c|}
		\hline
		\multirow{2}{*}{\textbf{Method}} &
		\multirow{2}{*}{\textbf{Backbone}} &
		\multirow{2}{*}{\textbf{Pre-train}} & \multicolumn{4}{c|}{\textbf{Macro}} & \multicolumn{4}{c|}{\textbf{Micro}} \\ \cline{4-11}
		& & & \textbf{Split 1} & \textbf{Split 2} & \textbf{Split 3} &  \textbf{Avg}& \textbf{Split 1} & \textbf{Split 2} & \textbf{Split 3} &  \textbf{Avg} \\ \hline \hline
		TSN~\cite{tsn}$^{\dagger\dagger}$ & R2D-34 & ImNet & 58.01 & 55.01 & 54.78 & 55.93 & 46.76 & 45.22 & 45.56 & 45.85 \\ \hline
		I3D~\cite{carreira2017quo}$^\dagger$ & I3D-IncV1 & Kin & - & - & - & - & 49.79 & - & - & - \\ \hline
		Li \etal~\cite{li2018eye}  & I3D-IncV1 & Kin & - & - & - & - & 53.30 & - & - & - \\ \hline
		Ego-RNN~\cite{egornn} & R2D-34 & ImNet & 62.17 & 61.47 & 58.63 & 60.76 & 53.84 & 52.37 & 50.85 & 52.35 \\ \hline
		MCN~\cite{huang2019mutual} & R3D-50-NL & Kin & 62.6 & - & - & - & 55.7 & - & - & \\ \hline
		RU-LSTM~\cite{rulstm} & BNInc$^\ddagger$ & ImNet & - & - & - & 60.20 & - & - & - & \\ \hline
		LSTA & R2D-34 & ImNet & 64.24 & 62.86 & 58.49 & 61.86 & 54.79 & 55.36 & 48.89 & 53.01 \\ \hline
		Lu \etal~\cite{lu2019learning} & I3D-IncV1 & Kin & 68.60 & 65.33 & 63.98 & 65.97 & 60.54 & 55.21 & 55.32 & 57.02 \\ \hline \hline
		I3D~\cite{carreira2017quo}$^\dagger$ & I3D-IncV1 & Kin & - & - & - & - & 47.26 & - & - & - \\ \hline
		eleGAtt~\cite{zhang2019eleatt}$^{\dagger\dagger}$ & R2D-34 & ImNet & 59.1 & 57.07 & 54.87 & 57.01 & 51.26 & 45.97 & 46.41 & 47.88 \\ \hline
		SAP~\cite{sap} & R3D-50$^\ddagger$ & Kin & 64.1 & 62.1 & 62.0 & 62.7 & - & - & - & - \\ \hline \hline
		EgoACO & R2D-101 & ImNet+EPIC & 68.69 & 68.45 & 63.78 & 66.97 & 61.6 & 59.56 & 56.33 & 59.16 \\ \hline
	\end{tabular}
	\label{tab:egtea_sota}
\end{table*}

%% file: sub/5_conclusions.tex
We presented EgoACO, an egocentric action recognition framework leveraging self-attention dictionaries in a rank-1 approximation of second order pooling to encode video frame features into object, context, action descriptors for clip classification. For the action descriptors we designed a recurrent encoding scheme based on it, LSTA, that extends LSTM with built-in attention on the input and on the memory propagation. LSTA is a general sequence learning model that can be plugged into RNN based video architectures. Self-attention endows EgoACO with built-in visual explanation that aid learning from video-level labels, and help interpreting predictions. We performed a detailed analysis of EgoACO on two large scale egocentric action recognition benchmarks, EPIC-KITCHENS and EGTEA Gaze+. Compared to other methods that use large scale pre-training, multi-modal input or extra annotations such as gaze, hand masks, object detections, \etc, EgoACO achieves competitive or state-of-the-art recognition performance, showing its effectiveness in addressing egocentric action recognition.